\documentclass{article} %
\usepackage{iclr2027_conference,times}
\usepackage{amsmath,amsfonts,bm}

\def\eqref#1{equation~\ref{#1}}

\def\1{\bm{1}}

\def\mA{{\bm{A}}}
\def\mB{{\bm{B}}}
\def\mC{{\bm{C}}}

\DeclareMathAlphabet{\mathsfit}{\encodingdefault}{\sfdefault}{m}{sl}
\SetMathAlphabet{\mathsfit}{bold}{\encodingdefault}{\sfdefault}{bx}{n}

\input{notation/probe-styles.tex}
\input{notation/probes.tex}

\newcommand{\effectiveAnswerSet}{\mathcal E}
\newcommand{\effectiveAnswer}[2]{e_{#1}(#2)}

\newcommand{\fineTuningSubjectSet}{T}
\newcommand{\heldOutSubjectSet}{H}

\newcommand{\setSize}[1]{\lvert #1 \rvert}
\newcommand{\labelPool}{\mathcal A}

\newcommand{\MiniCpmModel}{\mbox{MiniCPM5-2B-Base}}

\newcommand{\fttemplates}{fine-tuning templates}

\newcommand{\evaltemplate}{held-out evaluation template}

\newcommand{\sourcetemplate}{query template}

\newcommand{\neighbourPruning}{neighbor pruning}

\newcommand{\RelSetNameBear}{BEAR}
\newcommand{\RelSetNameEngineered}{Town}
\newcommand{\RelSetNameHernandez}{Hernandez}

\newcommand{\nHernandezRelations}{11}
\newcommand{\nHernandezRelationsWord}{eleven}
\newcommand{\nEngineeredRelations}{6}
\newcommand{\nEngineeredRelationsWord}{six}
\newcommand{\nBearRelations}{53}
\newcommand{\nBearRelationsWord}{fifty-three}
\newcommand{\nRosterModels}{11}
\newcommand{\nRosterModelsWord}{eleven}

\newcommand{\nPooledRelations}{\the\numexpr\nHernandezRelations+\nEngineeredRelations+\nBearRelations\relax}

\newcommand{\productionPoolSize}{179}
\newcommand{\poolEligibilityFloor}{86}

\newcommand{\nEligibleCategoryPoolsWord}{thirteen}
\newcommand{\poolZipfFloor}{2.32}
\newcommand{\poolCosineCeiling}{0.446}

\newcommand{\pruneStepsWord}{four}

\usepackage{hyperref}
\usepackage{url}
\usepackage{booktabs}
\usepackage[table]{xcolor} %
\usepackage{tikz} %
\usepackage{enumitem}
\usetikzlibrary{decorations.pathreplacing,arrows.meta,calc,shapes.misc}
\usepackage{graphicx}
\usepackage{caption} %
\usepackage{subcaption} %
\usepackage{placeins} %
\usepackage[capitalize,noabbrev]{cleveref} %
\crefformat{section}{\S#2#1#3}
\Crefformat{section}{Section~#2#1#3}

\title{{\raggedright
Knowledge That Is Linear Before\\
Fine-Tuning Generalizes After It\par}}

\author{Faridoun Mehri \& Hinrich Sch\"{u}tze \\
Center for Information and Language Processing, LMU Munich \\
Munich Center for Machine Learning (MCML) \\
\texttt{feraidoon@cis.lmu.de}
}

\newcounter{notecounter}
\newcommand{\enotesoff}{\long\gdef\enote##1##2{}}
\newcommand{\enoteson}{\long\gdef\enote##1##2{{
\stepcounter{notecounter}
{\large\bf
\hspace{0cm}\arabic{notecounter} $<<<$ ##1: ##2
$>>>$\hspace{1cm}}}}}

\enoteson
\enotesoff

  \iclrfinalcopy
  \hypersetup{pdftitle={Knowledge That Is Linear Before Fine-Tuning Generalizes After It},
    pdfauthor={Faridoun Mehri, Hinrich Sch\"utze}}
\ificlrfinal \usepackage[T1]{fontenc}\fi
\begin{document}

\maketitle
\lhead{}\renewcommand{\headrulewidth}{0pt}
\newcommand{\tableOneRawRMin}{0.87}
\newcommand{\tableOneRawRMax}{0.92}
\newcommand{\tableOneRawRMean}{0.89}
\newcommand{\tableOneRawRMedian}{0.90}
\newcommand{\tableOneRawRLowestBound}{0.78}
\newcommand{\tableOneRawRhoMin}{0.85}
\newcommand{\tableOneRawRhoMax}{0.90}
\newcommand{\tableOneRawRhoMean}{0.88}
\newcommand{\tableOneRawRhoMedian}{0.89}
\newcommand{\tableOneRawRhoLowestBound}{0.72}
\newcommand{\tableOneRawLowestBound}{0.72}
\newcommand{\tableOneRecallRMin}{0.79}
\newcommand{\tableOneRecallRMax}{0.86}
\newcommand{\tableOneRecallRMean}{0.84}
\newcommand{\tableOneRecallRMedian}{0.85}
\newcommand{\tableOneRecallRLowestBound}{0.67}
\newcommand{\tableOneRecallRhoMin}{0.79}
\newcommand{\tableOneRecallRhoMax}{0.87}
\newcommand{\tableOneRecallRhoMean}{0.84}
\newcommand{\tableOneRecallRhoMedian}{0.84}
\newcommand{\tableOneRecallRhoLowestBound}{0.64}
\newcommand{\tableOneRecallLowestBound}{0.64}
\newcommand{\tableOneAnswerCountRMin}{0.77}
\newcommand{\tableOneAnswerCountRMax}{0.88}
\newcommand{\tableOneAnswerCountRMean}{0.83}
\newcommand{\tableOneAnswerCountRMedian}{0.84}
\newcommand{\tableOneAnswerCountRLowestBound}{0.64}
\newcommand{\tableOneAnswerCountRhoMin}{0.81}
\newcommand{\tableOneAnswerCountRhoMax}{0.88}
\newcommand{\tableOneAnswerCountRhoMean}{0.86}
\newcommand{\tableOneAnswerCountRhoMedian}{0.86}
\newcommand{\tableOneAnswerCountRhoLowestBound}{0.64}
\newcommand{\tableOneAnswerCountLowestBound}{0.64}
\newcommand{\tableOneSubjectCountRMin}{0.82}
\newcommand{\tableOneSubjectCountRMax}{0.89}
\newcommand{\tableOneSubjectCountRMean}{0.86}
\newcommand{\tableOneSubjectCountRMedian}{0.87}
\newcommand{\tableOneSubjectCountRLowestBound}{0.71}
\newcommand{\tableOneSubjectCountRhoMin}{0.84}
\newcommand{\tableOneSubjectCountRhoMax}{0.89}
\newcommand{\tableOneSubjectCountRhoMean}{0.88}
\newcommand{\tableOneSubjectCountRhoMedian}{0.88}
\newcommand{\tableOneSubjectCountRhoLowestBound}{0.71}
\newcommand{\tableOneSubjectCountLowestBound}{0.71}
\newcommand{\tableOneSubjectFrequencyRMin}{0.85}
\newcommand{\tableOneSubjectFrequencyRMax}{0.91}
\newcommand{\tableOneSubjectFrequencyRMean}{0.88}
\newcommand{\tableOneSubjectFrequencyRMedian}{0.88}
\newcommand{\tableOneSubjectFrequencyRLowestBound}{0.74}
\newcommand{\tableOneSubjectFrequencyRhoMin}{0.85}
\newcommand{\tableOneSubjectFrequencyRhoMax}{0.90}
\newcommand{\tableOneSubjectFrequencyRhoMean}{0.88}
\newcommand{\tableOneSubjectFrequencyRhoMedian}{0.88}
\newcommand{\tableOneSubjectFrequencyRhoLowestBound}{0.73}
\newcommand{\tableOneSubjectFrequencyLowestBound}{0.73}
\newcommand{\tableOneAllJointlyRMin}{0.70}
\newcommand{\tableOneAllJointlyRMax}{0.82}
\newcommand{\tableOneAllJointlyRMean}{0.77}
\newcommand{\tableOneAllJointlyRMedian}{0.77}
\newcommand{\tableOneAllJointlyRLowestBound}{0.56}
\newcommand{\tableOneAllJointlyRhoMin}{0.75}
\newcommand{\tableOneAllJointlyRhoMax}{0.85}
\newcommand{\tableOneAllJointlyRhoMean}{0.82}
\newcommand{\tableOneAllJointlyRhoMedian}{0.83}
\newcommand{\tableOneAllJointlyRhoLowestBound}{0.56}
\newcommand{\tableOneAllJointlyLowestBound}{0.56}
\newcommand{\tableOneSingleControlRMin}{0.77}
\newcommand{\tableOneSingleControlRMax}{0.91}
\newcommand{\tableOneSingleControlRhoMin}{0.79}
\newcommand{\tableOneSingleControlRhoMax}{0.90}

\newcommand{\labelSetPools}{13}
\newcommand{\labelSetPoolsWord}{thirteen}
\newcommand{\labelSetRelations}{70}
\newcommand{\labelSetDroppedRelations}{0}
\newcommand{\labelSetReferenceR}{0.92}
\newcommand{\labelSetRMin}{0.91}
\newcommand{\labelSetRMax}{0.93}
\newcommand{\labelSetRMedian}{0.92}
\newcommand{\labelSetRMean}{0.92}
\newcommand{\labelSetRLowestBound}{0.85}
\newcommand{\labelSetRExcludesZero}{13}
\newcommand{\labelSetRExcludesZeroWord}{thirteen}
\newcommand{\labelSetRankRhoMin}{0.95}
\newcommand{\labelSetRankRhoMax}{0.97}

\newcommand{\repVarHernandezR}{0.97}
\newcommand{\repVarHernandezRLo}{0.93}
\newcommand{\repVarHernandezRHi}{0.98}
\newcommand{\repVarHernandezCells}{110}
\newcommand{\repVarHernandezShareMin}{1.3}
\newcommand{\repVarHernandezShareMax}{10.3}
\newcommand{\repVarHernandezShareMaxModel}{Granite-4.2-3B}
\newcommand{\repVarHernandezShareMean}{3.5}
\newcommand{\repVarHernandezShareMedian}{2.8}
\newcommand{\repVarHernandezShareCiHiMax}{27.5}
\newcommand{\repVarEngineeredR}{0.93}
\newcommand{\repVarEngineeredRLo}{0.79}
\newcommand{\repVarEngineeredRHi}{0.95}
\newcommand{\repVarEngineeredCells}{60}
\newcommand{\repVarEngineeredShareMin}{2.7}
\newcommand{\repVarEngineeredShareMax}{11.5}
\newcommand{\repVarEngineeredShareMaxModel}{G9v3-3B}
\newcommand{\repVarEngineeredShareMean}{6.4}
\newcommand{\repVarEngineeredShareMedian}{5.9}
\newcommand{\repVarEngineeredShareCiHiMax}{81.4}
\newcommand{\repVarBearR}{0.96}
\newcommand{\repVarBearRLo}{0.92}
\newcommand{\repVarBearRHi}{0.98}
\newcommand{\repVarBearCells}{53}
\newcommand{\repVarBearShare}{3.9}
\newcommand{\repVarBearShareLo}{2.2}
\newcommand{\repVarBearShareHi}{8.0}

\newcommand{\pairCoveredRelations}{55}
\newcommand{\pairCoveredHernandez}{11}
\newcommand{\pairPanelHernandez}{11}
\newcommand{\pairCoveredTown}{2}
\newcommand{\pairPanelTown}{6}
\newcommand{\pairCoveredBear}{42}
\newcommand{\pairPanelBear}{53}
\newcommand{\pairRawRMin}{0.85}
\newcommand{\pairRawRMax}{0.91}
\newcommand{\pairAloneRMin}{0.82}
\newcommand{\pairAloneRMax}{0.90}
\newcommand{\pairAloneRMedian}{0.85}
\newcommand{\pairAloneRMean}{0.86}
\newcommand{\pairAloneRLowestBound}{0.71}
\newcommand{\pairAloneRhoMin}{0.81}
\newcommand{\pairAloneRhoMax}{0.91}
\newcommand{\pairSubjRMin}{0.82}
\newcommand{\pairSubjRMax}{0.90}
\newcommand{\pairAllMaxShift}{0.02}

\newcommand{\predictorGainRMin}{0.87}
\newcommand{\predictorGainRMax}{0.92}
\newcommand{\predictorGainRMean}{0.89}
\newcommand{\predictorGainRMedian}{0.90}
\newcommand{\predictorGainRExcludesZero}{11}
\newcommand{\predictorGainRExcludesZeroWord}{eleven}
\newcommand{\predictorGainRhoMin}{0.85}
\newcommand{\predictorGainRhoMax}{0.90}
\newcommand{\predictorGainRhoMean}{0.88}
\newcommand{\predictorGainRhoMedian}{0.89}
\newcommand{\predictorGainRhoExcludesZero}{11}
\newcommand{\predictorGainRhoExcludesZeroWord}{eleven}
\newcommand{\predictorGainGivenRecallRMin}{0.79}
\newcommand{\predictorGainGivenRecallRMax}{0.86}
\newcommand{\predictorGainGivenRecallRMean}{0.84}
\newcommand{\predictorGainGivenRecallRMedian}{0.85}
\newcommand{\predictorGainGivenRecallRExcludesZero}{11}
\newcommand{\predictorGainGivenRecallRExcludesZeroWord}{eleven}
\newcommand{\predictorGainGivenRecallRhoMin}{0.79}
\newcommand{\predictorGainGivenRecallRhoMax}{0.87}
\newcommand{\predictorGainGivenRecallRhoMean}{0.84}
\newcommand{\predictorGainGivenRecallRhoMedian}{0.84}
\newcommand{\predictorGainGivenRecallRhoExcludesZero}{11}
\newcommand{\predictorGainGivenRecallRhoExcludesZeroWord}{eleven}
\newcommand{\predictorRecallRMin}{0.63}
\newcommand{\predictorRecallRMax}{0.74}
\newcommand{\predictorRecallRMean}{0.69}
\newcommand{\predictorRecallRMedian}{0.69}
\newcommand{\predictorRecallRExcludesZero}{11}
\newcommand{\predictorRecallRExcludesZeroWord}{eleven}
\newcommand{\predictorRecallRhoMin}{0.50}
\newcommand{\predictorRecallRhoMax}{0.64}
\newcommand{\predictorRecallRhoMean}{0.56}
\newcommand{\predictorRecallRhoMedian}{0.57}
\newcommand{\predictorRecallRhoExcludesZero}{11}
\newcommand{\predictorRecallRhoExcludesZeroWord}{eleven}
\newcommand{\predictorRecallGivenGainRMin}{0.40}
\newcommand{\predictorRecallGivenGainRMax}{0.58}
\newcommand{\predictorRecallGivenGainRMean}{0.46}
\newcommand{\predictorRecallGivenGainRMedian}{0.45}
\newcommand{\predictorRecallGivenGainRExcludesZero}{11}
\newcommand{\predictorRecallGivenGainRExcludesZeroWord}{eleven}
\newcommand{\predictorRecallGivenGainRhoMin}{0.25}
\newcommand{\predictorRecallGivenGainRhoMax}{0.42}
\newcommand{\predictorRecallGivenGainRhoMean}{0.33}
\newcommand{\predictorRecallGivenGainRhoMedian}{0.32}
\newcommand{\predictorRecallGivenGainRhoExcludesZero}{8}
\newcommand{\predictorRecallGivenGainRhoExcludesZeroWord}{eight}

\newcommand{\repeatCorrAllRelations}{70}
\newcommand{\repeatCorrSubsetRelations}{17}
\newcommand{\repeatCorrSubsetModelsWord}{ten}
\newcommand{\repeatMinicpmRRepeatZero}{0.92}
\newcommand{\repeatMinicpmRRepeatOne}{0.93}
\newcommand{\repeatMinicpmRRepeatTwo}{0.92}
\newcommand{\repeatMinicpmRMeanOfThree}{0.93}
\newcommand{\repeatMinicpmRhoRepeatZero}{0.90}
\newcommand{\repeatMinicpmRhoRepeatOne}{0.89}
\newcommand{\repeatMinicpmRhoRepeatTwo}{0.92}
\newcommand{\repeatMinicpmRhoMeanOfThree}{0.92}
\newcommand{\repeatSubsetRMin}{0.78}
\newcommand{\repeatSubsetRMax}{0.95}
\newcommand{\repeatSubsetRhoMin}{0.75}
\newcommand{\repeatSubsetRhoMax}{0.95}

\newcommand{\failErrors}{37{,}571}
\newcommand{\failCells}{757}
\newcommand{\failExplainedObserved}{48.3}
\newcommand{\failExplainedObservedH}{1.4}
\newcommand{\failExplainedShuffled}{26.7}
\newcommand{\failExplainedShuffledH}{1.4}
\newcommand{\failExplainedChance}{17.8}
\newcommand{\failExplainedModelObservedMin}{42.6}
\newcommand{\failExplainedModelObservedMax}{57.3}
\newcommand{\failExplainedModelShuffledMin}{24.4}
\newcommand{\failExplainedModelShuffledMax}{28.9}
\newcommand{\failFamilyProbeObserved}{21.5}
\newcommand{\failFamilyProbeShuffled}{8.8}
\newcommand{\failFamilyProbeChance}{5.6}
\newcommand{\failFamilyRecallObserved}{11.5}
\newcommand{\failFamilyRecallShuffled}{7.3}
\newcommand{\failFamilyRecallChance}{6.4}
\newcommand{\failFamilyFineTuningObserved}{12.9}
\newcommand{\failFamilyFineTuningShuffled}{8.3}
\newcommand{\failFamilyFineTuningChance}{3.3}
\newcommand{\failFamilyEpochZeroObserved}{2.4}
\newcommand{\failFamilyEpochZeroShuffled}{2.3}
\newcommand{\failFamilyEpochZeroChance}{2.5}
\newcommand{\failShareProbeRankOneObserved}{8.8}
\newcommand{\failShareProbeRankOneShuffled}{3.3}
\newcommand{\failShareProbeRankOneChance}{1.5}
\newcommand{\failShareProbeRankTwoObserved}{7.1}
\newcommand{\failShareProbeRankTwoShuffled}{2.8}
\newcommand{\failShareProbeRankTwoChance}{1.9}
\newcommand{\failShareProbeRankThreeObserved}{5.6}
\newcommand{\failShareProbeRankThreeShuffled}{2.7}
\newcommand{\failShareProbeRankThreeChance}{2.2}
\newcommand{\failShareRecallRankOneObserved}{4.0}
\newcommand{\failShareRecallRankOneShuffled}{2.6}
\newcommand{\failShareRecallRankOneChance}{1.9}
\newcommand{\failShareRecallRankTwoObserved}{4.0}
\newcommand{\failShareRecallRankTwoShuffled}{2.3}
\newcommand{\failShareRecallRankTwoChance}{2.2}
\newcommand{\failShareRecallRankThreeObserved}{3.5}
\newcommand{\failShareRecallRankThreeShuffled}{2.4}
\newcommand{\failShareRecallRankThreeChance}{2.3}
\newcommand{\failShareMostFrequentObserved}{5.0}
\newcommand{\failShareMostFrequentShuffled}{6.1}
\newcommand{\failShareMostFrequentChance}{1.9}
\newcommand{\failShareNearestObserved}{8.0}
\newcommand{\failShareNearestShuffled}{2.3}
\newcommand{\failShareNearestChance}{1.4}
\newcommand{\failShareEpochZeroObserved}{2.4}
\newcommand{\failShareEpochZeroShuffled}{2.3}
\newcommand{\failShareEpochZeroChance}{2.5}

\begin{abstract}
  Not all facts a language model can recall support further learning equally well. We distinguish \emph{active} facts, which support generalization beyond their own recall, from \emph{inert} facts, which can be recalled but do not support such generalization in a given task. We study this distinction by constructing new relations from existing ones.
  Take the relation that maps a place to its continent (Paris $\to$ Europe) and assign each of its answers an arbitrary new label (Europe $\to$ crab, Asia $\to$ zebra).
  The new relation then maps Paris $\to$ crab.
  We fine-tune the model on the new relation for a subset of subjects, without telling it the original relation or the relabeling rule, and test whether it assigns the corresponding labels to held-out subjects that share an answer in the original relation. Having learned Paris $\to$ crab, does it answer Munich $\to$ crab?
  Across \nRosterModelsWord{} language models and
70
  relations, how linearly readable the original relation is in pretrained representations strongly predicts this generalization. Within each model, the two correlate across relations at Pearson $r$ = $\tableOneRawRMin$ to $\tableOneRawRMax$. This association persists when controlling, separately and jointly, for the number of fine-tuning subjects and answers, the model's accuracy on the original relation, and how often the subjects occur in pretraining text.
  It is also robust to the arbitrary choices in the construction: re-randomizing the label assignment and fine-tuning seed, or replacing the animal labels with labels from \nEligibleCategoryPoolsWord{} other semantic categories, leaves it essentially unchanged.
  When we map a fine-tuned model's wrong label back to the original answer it was assigned to, Munich $\to$ zebra becomes Munich $\to$ Asia. Answers recovered this way are more often among the linear probe's top answers for the subject than among the pretrained model's own.
  These findings tie the active/inert distinction to linear readability. Whether a new relation generalizes depends less on whether its facts can be recalled than on how readily their shared structure can be read out linearly.
\end{abstract}

\begin{figure}[t]
  \centering
  \expandafter\ifx\csname ifFigSymbols\endcsname\relax \expandafter\newif\csname ifFigSymbols\endcsname\fi
\newcommand{\cgEight}{\fontsize{8}{9.6}\selectfont}
\definecolor{cgInk}{HTML}{30343B}
\definecolor{cgGray}{HTML}{7A828E}
\definecolor{cgPanel}{HTML}{F4F6F8}
\definecolor{cgPanelLine}{HTML}{D9DDE2}
\definecolor{cgRule}{HTML}{E3E6EA}
\definecolor{cgTeal}{HTML}{2A9D8F}
\definecolor{cgPlum}{HTML}{8C5BA8}
\definecolor{cgRose}{HTML}{D0607A}
\definecolor{cgBlue}{HTML}{3F6FA8}
\definecolor{cgOrange}{HTML}{E08A1E}
\definecolor{cgGood}{HTML}{2E7D4F}
\definecolor{cgBad}{HTML}{B23A3A}
\newcommand{\cgQueryA}{\ldots\unkern\ The medallion of}
\newcommand{\cgQueryB}{features the}
\newlength{\cgQAW}\settowidth{\cgQAW}{\cgEight\cgQueryA}
\newlength{\cgQBW}\settowidth{\cgQBW}{\cgEight\cgQueryB}
\newlength{\cgWordW}\newlength{\cgTmpW}\newlength{\cgPiW}\settowidth{\cgPiW}{\cgEight$\pi$}
\foreach \w in {crab, zebra, boar, panda, robin, snail} {%
  \settowidth{\cgTmpW}{\cgEight\itshape \w}\ifdim\cgTmpW>\cgWordW\global\cgWordW=\cgTmpW\fi}
\newlength{\cgNameW}
\foreach \w in {\termRecallShort{}, \termGainShort{}, \newrelgen{}} {%
  \settowidth{\cgTmpW}{\cgEight\w}\ifdim\cgTmpW>\cgNameW\global\cgNameW=\cgTmpW\fi}
\def\cgDrop{0.27}\def\cgTail{0.06}\def\cgGap{0.13}
\newcommand{\cgStep}[1]{\tikz[baseline=-2.65pt]{\fill[cgInk] (0,0) circle (0.12);
    \node[anchor=center, text=white, font=\scriptsize\mdseries, inner sep=0pt] at (0,0) {#1};}\hspace{0.4em}}
\newcommand{\cgMeasure}[5]{%
  \node[mname] at (\xr,\cgY) {#1};
  \ifnum#5=1 \node[gloss] at (\xr,\cgY-\cgDrop) {#4};\fi
  \pgfmathsetmacro{\u}{\cgY-\cgDrop}%
  \ifnum#3=1
    \node[pill, fill=#2, draw=#2, text=white] at (\xt,\cgY+0.09) {high};
  \else
    \node[pill, fill=white, draw=#2, text=#2] at (\xt,\cgY+0.09) {low};
  \fi
  \pgfmathsetmacro{\cgYnext}{\u-\cgTail-\cgGap-0.19}\global\let\cgY\cgYnext
}
\begin{tikzpicture}[x=1cm, y=1cm, font=\cgEight, text=cgInk, inner sep=0pt,
    panel/.style={draw=cgPanelLine, line width=0.4pt, fill=cgPanel, rounded corners=3pt},
    head/.style={anchor=base, font=\cgEight\bfseries},
    subhead/.style={anchor=base, font=\scriptsize, text=cgGray},
    ray/.style={draw=cgGray, line width=0.5pt},
    flow/.style={-{Stealth[length=3.5pt, width=3pt]}, line width=0.5pt, draw=cgInk},
    maparr/.style={-{Stealth[length=3pt]}, line width=0.5pt, draw=cgInk},
    ansT/.style={fill=cgTeal}, ansP/.style={fill=cgPlum}, ansR/.style={fill=cgRose},
    held/.style={draw=cgTeal, line width=0.9pt, fill=white},
    chip/.style={draw=cgOrange, line width=0.6pt, fill=white, rounded corners=2.5pt,
                 minimum width=0.85cm, minimum height=0.45cm, font=\cgEight\itshape},
    verdict/.style={line width=1.2pt, line cap=round, line join=round},
    mname/.style={anchor=base west, font=\cgEight},
    gloss/.style={anchor=base west, font=\scriptsize, text=cgGray},
    pill/.style={rounded rectangle, minimum width=0.62cm, minimum height=0.28cm, line width=0.5pt,
                 font=\scriptsize, anchor=west},
    swatch/.style={baseline=-0.55ex}]

  \def\mA{3.42}  \def\cAw{2.8}     %
  \def\mB{6.42}  \def\cBw{2.45}    %
  \def\mC{9.51}  \def\cCw{3.05}    %
  \def\xr{11.38} \def\xe{13.91}    %
  \def\ph{2.4}                     %
  \pgfmathsetmacro{\cA}{\mA-\cAw/2}
  \pgfmathsetmacro{\cB}{\mB-\cBw/2}
  \pgfmathsetmacro{\cC}{\mC-\cCw/2}
  \pgfmathsetmacro{\xt}{\xr + \cgNameW/1cm + 0.14}   %

  \foreach \x/\n/\one/\two/\three in {
      \mA/1/Pretrained model/its internal representation/{of each subject, drawn in 2-D},
      \mB/2/Relabel the answers/{\ifFigSymbols a random bijection $\pi$\else a random bijection maps\fi}/{\ifFigSymbols maps answers to animals\else each answer to an animal\fi},
      \mC/3/{Fine-tune, then test}/{\ifFigSymbols train on subjects in $\fineTuningSubjectSet$, then ask\else train on most subjects,\fi}/{\ifFigSymbols about a held-out one ($\heldOutSubjectSet$)\else test on a held-out one\fi}} {
    \node[head] at (\x,-0.23) {\cgStep{\n}\one};
    \node[subhead] at (\x,-0.49) {\two};
    \node[subhead] at (\x,-0.75) {\three};
  }
  \node[head, anchor=base west] at (\xr,-0.23) {What we measure};

  \draw[draw=cgRule, line width=0.4pt] (0.05,-3.63) -- (\xe,-3.63);

  \foreach \bot/\name/\la/\lb/\lc in {-3.42/A/active fact:/recalled and/generalizes,
                                      -6.24/B/inert fact:/recalled but/does not generalize} {
    \pgfmathsetmacro{\mid}{\bot+\ph/2}
    \node[anchor=west, align=left, font=\scriptsize\itshape, text=cgGray] at (0.01,\mid)
      {{\cgEight\bfseries\upshape\color{cgInk}Relation \name}\\[1.5pt] \la\\ \lb\\ \lc};
    \draw[panel] (\cA,\bot) rectangle ++(\cAw,\ph);
    \draw[panel] (\cB,\bot) rectangle ++(\cBw,\ph);
    \draw[panel] (\cC,\bot) rectangle ++(\cCw,\ph);
    \draw[flow] (\cA+\cAw+0.04,\mid) -- (\cB-0.04,\mid);
    \draw[flow] (\cB+\cBw+0.04,\mid) -- (\cC-0.04,\mid);
    \begin{scope}[shift={(\cA,\bot)}]
      \draw[ray] (1.4,\ph-0.12) -- (1.4,0.55*\ph) -- (0.25,0.2*\ph);
      \draw[ray] (1.4,0.55*\ph) -- (2.55,0.2*\ph);
    \end{scope}
  }

  \begin{scope}[shift={(\cA,-3.42)}]
    \foreach \p in {(0.40,1.95), (0.68,2.11), (0.95,1.92), (0.30,1.61), (0.92,1.56), (0.42,1.29), (0.78,1.27)}
      \fill[ansT] \p circle (0.07);
    \foreach \p in {(2.20,2.08), (2.48,1.92), (1.90,1.83), (2.22,1.72), (2.52,1.58), (1.84,1.47), (2.16,1.36)}
      \fill[ansP] \p circle (0.07);
    \foreach \p in {(1.12,0.70), (1.40,0.88), (1.68,0.68), (0.98,0.38), (1.30,0.45), (1.60,0.34), (1.88,0.45)}
      \fill[ansR] \p circle (0.07);
    \draw[held] (0.64,1.65) circle (0.08);
  \end{scope}

  \begin{scope}[shift={(\cA,-6.24)}]
    \foreach \p in {(0.57,1.25), (1.86,0.59), (2.34,1.74), (0.84,0.23), (2.23,1.23), (1.16,1.60),
                    (1.73,1.54), (0.50,0.31), (0.60,2.14), (1.65,2.09), (0.19,1.54)}
      \fill[ansT] \p circle (0.07);
    \foreach \p in {(2.02,1.96), (0.46,1.82), (0.24,1.12), (2.28,0.37), (2.55,2.04), (1.54,0.67),
                    (0.91,2.03), (1.68,0.26), (0.76,0.55), (1.92,1.23), (0.97,1.29)}
      \fill[ansP] \p circle (0.07);
    \foreach \p in {(0.76,1.54), (0.21,2.03), (2.60,0.82), (1.38,0.33), (1.37,0.96), (2.52,1.44),
                    (1.99,0.24), (2.27,2.17), (0.21,0.77), (2.07,1.58), (1.18,2.18)}
      \fill[ansR] \p circle (0.07);
    \draw[held] (1.09,0.71) circle (0.08);
  \end{scope}

  \pgfmathsetmacro{\wmax}{\cgWordW/1cm}
  \pgfmathsetmacro{\piw}{\cgPiW/1cm}
  \pgfmathsetmacro{\blockw}{0.1 + 0.72 + \wmax + \ifFigSymbols 0.1 + 0.22 + \piw\else 0\fi}
  \pgfmathsetmacro{\xzero}{\mB - \blockw/2 + 0.1}
  \pgfmathsetmacro{\xbrace}{\xzero + 0.72 + \wmax + 0.1}
  \foreach \mid/\wa/\wb/\wc in {-2.22/crab/zebra/boar, -5.04/panda/robin/snail} {
    \foreach \k/\c/\w in {0/ansT/\wa, 1/ansP/\wb, 2/ansR/\wc} {
      \pgfmathsetmacro{\yc}{\mid + 0.508 - 0.508*\k}
      \fill[\c] (\xzero,\yc) circle (0.1);
      \draw[maparr] (\xzero+0.17,\yc) -- ++(0.45,0);
      \node[anchor=base west, font=\cgEight\itshape] at (\xzero+0.72,\yc-0.065) {\w};
    }
    \ifFigSymbols
      \draw[draw=cgInk, line width=0.5pt, decorate, decoration={brace, amplitude=3pt}]
        (\xbrace,\mid+0.66) -- (\xbrace,\mid-0.66);
      \node[anchor=west] at (\xbrace+0.22,\mid) {$\pi$};
    \fi
  }

  \pgfmathsetmacro{\xq}{\mC - max(\cgQAW/1cm + 0.25, \cgQBW/1cm + 0.08 + 0.85 + 0.42)/2}
  \foreach \mid/\ans/\row in {-2.22/crab/a, -5.04/snail/b} {
    \node[anchor=base west, font=\scriptsize, text=cgGray] at (\xq,\mid+0.47) {new relation's test prompt:};
    \node[anchor=base west] (q1\row) at (\xq,\mid+0.07) {\cgQueryA};
    \draw[held] ([xshift=0.18cm]q1\row.base east) ++(0,0.065) circle (0.08);
    \node[anchor=base west] (q2\row) at (\xq,\mid-0.45) {\cgQueryB};
    \node[chip, anchor=base west] (chip\row) at ([xshift=0.08cm]q2\row.base east) {\ans};
  }
  \draw[verdict, draw=cgGood] ([xshift=0.13cm]chipa.east) -- ++(0.09,-0.11) -- ++(0.18,0.27);
  \draw[verdict, draw=cgBad] ([xshift=0.15cm]chipb.east) ++(0,0.13) -- ++(0.24,-0.26)
                          ([xshift=0.15cm]chipb.east) ++(0,-0.13) -- ++(0.24,0.26);
  \node[anchor=base, font=\scriptsize, text=cgGray] at ([yshift=-0.3cm]chipb.south) {expected:\ panda};

  \foreach \mid/\withgloss/\gain/\gen in {-2.22/1/1/1, -5.04/0/0/0} {
    \pgfmathsetmacro{\cgY}{\mid + (0.19 + 3*\cgDrop + 3*\cgTail + 2*(\cgGap+0.19))/2 - 0.19}
    \cgMeasure{\termRecallShort{}}{cgBlue}{1}{pretrained model's acc.}{\withgloss}
    \cgMeasure{\termGainShort{}}{cgBlue}{\gain}{probe gain over chance}{\withgloss}
    \cgMeasure{\newrelgen{}}{cgOrange}{\gen}{held-out subject accuracy}{\withgloss}
  }

  \node[anchor=base, font=\scriptsize, text=cgGray] at (\xe/2,-6.66) {%
    \tikz[swatch]\fill[cgTeal] (0,0) circle (0.07);\kern3pt subject used for fine-tuning\ifFigSymbols\ ($\fineTuningSubjectSet$)\fi\hspace{0.4cm}%
    \tikz[swatch]\draw[cgTeal, line width=0.9pt, fill=white] (0,0) circle (0.08);\kern3pt held-out subject, not fine-tuned or probed\ifFigSymbols\ ($\heldOutSubjectSet$)\fi};
  \node[anchor=base, font=\scriptsize, text=cgGray] at (\xe/2,-6.93) {%
    \tikz[swatch]{\fill[cgTeal] (0,0) circle (0.07); \fill[cgPlum] (0.2,0) circle (0.07); \fill[cgRose] (0.4,0) circle (0.07);}\kern3pt color: the subject's original answer\hspace{0.4cm}%
    \tikz[swatch]\draw[cgGray, line width=0.5pt] (0,0) -- (0.45,0);\kern3pt linear probe's boundaries between the answers};
\end{tikzpicture}
  \caption{\textbf{Facts generalize to a new relation when they are linearly readable.} The
    pretrained model recalls relations A and B equally well, but only A's answers are linearly
    readable from its representations of the subjects. A random bijection from the answers to animal
    labels defines a \newrel{}. After fine-tuning on it, the model generalizes A's labels to held-out
    subjects but not B's. The fine-tuned model's wrong label for B's held-out subject, ``snail,'' is the label
    of the answer the linear probe picks for that subject, not of its true answer (\hyperref[par:results-failures]{\newrelgen{} failures}). \termGainShort{} predicts \newrelgen{} better than \termRecallShort{} does
    (\hyperref[par:results-predictors]{\termGainShort{} versus \termRecallShort{}}).
    \Cref{sec:method} defines the three measures.}
  \label{fig:readable-tangled}
\end{figure}

\section{Introduction}
\label{sec:introduction}

Imagine that the pretraining text describes Mary as having grown up in France,
attending French schools, writing books in French and working in French-speaking
institutions, but never states her native language. A language model should
still assume that Mary is a native speaker of French. We call this
\emph{statistical generalization of facts}. The model infers a new fact from
facts it knows, by a statistical rule that the pretraining text rarely states
explicitly and that is not always true. It differs from entailment, where the new
fact follows necessarily from the known ones, and from paraphrase, where the
same fact is merely worded differently (\cref{sec:related-work}). Statistical
generalization is an important part of what makes language models useful beyond
recalling what they were told. Prior work has asked whether models draw such
inferences from facts they are trained on
\citep{DBLP:journals/corr/abs-2412-04614,treutlein2024connecting},
but, to our
knowledge, no prior work  has asked \emph{which} of a model's facts support them.

Not all facts a model knows support such generalization equally well. We
distinguish \emph{active} facts, which support generalization beyond their own
recall, from \emph{inert} facts, which the model can recall but which do not
support such generalization in a given task. The distinction
is important 
because 
a model may state
a fact reliably and still be unable to use it when learning
something new.
Thus, our research question is:
\emph{What property of how a pretrained model
encodes its knowledge determines whether new knowledge built on it
generalizes?}

We hypothesize that the answer is linear structure. Linear relational
embeddings were originally proposed as a mechanism for generalizing to unseen
relational facts \citep{DBLP:conf/icml/PaccanaroH00}, and many relations in
modern language models are, to a good approximation, decoded linearly from the
subject's representation \citep{hernandez2024linearity,merullo2025linear}. In a
synthetic setting, knowledge that is linearly encoded on the entity is the
knowledge that fine-tuning can extract \citep{allenzhu2024physics}. But
whether the linearity of a pretrained model's real-world knowledge affects how
\emph{new} knowledge built on it generalizes has not been
tested.
The hypothesis we investigate is that the more linearly readable a relation's
answers are in the pretrained model, which we call the relation's
\emph{\termGainShort{}}, the better new knowledge built on that relation
generalizes after fine-tuning.

To this end, we construct new relations from existing ones
(\cref{fig:readable-tangled}). Take the \prerel{} that maps a place to its
continent (Paris $\to$ Europe) and relabel each of its answers at random with
a different arbitrary word, such as an animal name (Europe $\to$ crab, Asia $\to$
zebra). The resulting \newrel{} maps Paris $\to$ crab. We fine-tune the model
on the \newrel{} for a subset of subjects, without telling it the original
relation or the relabeling rule, and test whether it assigns the right labels to
held-out subjects. Having learned Paris $\to$ crab, does it answer Munich
$\to$ crab? The design directly tests for statistical generalization. The \newrel{} is
unknown to the model,  but its structure is inherited from the
\prerel{}, so the model can generalize only by building on what it already
knows. We measure \termGainShort{} with a cross-validated linear probe on the
pretrained model's subject representations, \emph{\termRecallShort{}} as the
pretrained model's accuracy on the \prerel{}, and \emph{\newrelgen{}} as the
fine-tuned model's accuracy on held-out subjects
(\cref{sec:method}).

Across \nRosterModelsWord{} open-weight language models, base and
post-trained, from 2B to 27B parameters, and
70
relations, \termGainShort{} strongly predicts \newrelgen{}. Within each model,
the two correlate across relations at Pearson $r$ = \tableOneRawRMin{} to
\tableOneRawRMax{} (median \tableOneRawRMedian{}; \cref{sec:results}). The
association is not explained by how well the model recalls the relation, by
the numbers of answers and fine-tuning subjects, or by how often the subjects
occur in pretraining text. Controlling for all four jointly, $r$ is still
\tableOneAllJointlyRMin{} to \tableOneAllJointlyRMax{} (\cref{tab:cls-controls}).
So being active
is not the same as being known.
Finally, the probe also tells us how the models fail. When we map a fine-tuned
model's wrong label back to the original answer it was assigned to,
Munich $\to$ zebra becomes Munich $\to$ Asia. Answers recovered this way are more
often among the probe's top answers for the subject than among the pretrained
model's (\cref{fig:failure-shapley-main}).

Our contributions are:
\textbf{(1)} We introduce the distinction between active and inert facts, and a
  paradigm that measures it. The paradigm builds new relations by randomly relabeling the
  answers of existing ones.
\textbf{(2)}  We show, across \nRosterModelsWord{} models and
70
  relations, that how linearly readable a relation is before fine-tuning
  predicts how well a new relation built on it generalizes after fine-tuning.
\textbf{(3)}  We show that \termGainShort{} 
is a better predictor of \newrelgen{}
 than \termRecallShort{}, and that it still predicts \newrelgen{} when \termRecallShort{},
relation size and subject frequency are controlled for jointly.
Furthermore,
the probe predicts the fine-tuned models' errors better
  than the other explanations we test.

\section{Related work}
\label{sec:related-work}

Most related to our distinction between
inert vs active facts is work on inference from facts that
we will call ``entailment.'' A fact $f$ is entailed
by a set of other facts
$F$ if from accepting the truth of $F$, it
necessarily follows that the truth of $f$ has to be
accepted. For example,
\{``John lives in Tokyo'', ``Tokyo is in Japan'', ``People in Japan speak
Japanese''\} entails ``Where John lives, people speak Japanese''
\citep{DBLP:journals/corr/abs-2412-04614}. 
This type of entailment includes transitive inference,
inverse relations and kinship reasoning and it
has
been widely studied \citep[e.g.,][]{berglund2024the,mehrafarin-etal-2024-reasoning,efeoglu2026reversingarrowslargelanguage,DBLP:journals/corr/abs-2604-12176,DBLP:conf/acl/YangGKG024,Wang2024GrokkedTA,lampinen2025generalizationlanguagemodelsincontext}.
There is mixed evidence on how well LMs perform on
entailment
\citep{3495724.3497422,3491440.3491977}.

Statistical generalization (see
\cref{sec:introduction}) differs
from entailment in that the underlying statistical rule is
not 100\% accurate. Its accuracy is a matter of degree:
each of the facts given about Mary (in our earlier example) is additional
evidence and decreases the probability of Mary not being a
native French speaker.

Paraphrases, which can be viewed as linguistic
generalization, also have been widely studied
\citep[e.g.,][]{elazar2021consistency,jiang-etal-2020-know,chang2024largelanguagemodelsacquire,behnia2025factsstatsimpactspretraining,wang2025iclr-generalization}.
Similarly, there is work on the acquisition of linguistic
constructions if no or little evidence is present in the
pretraining corpus \citep[e.g.,][]{DBLP:conf/emnlp/MisraM24,10.1162/tacl_a_00720}.
Paraphrases and the acquisition of linguistic constructions
concern variation in how
the same underlying fact 
is expressed linguistically and are thus fundamentally different from statistical
generalization of facts.

Statistical generalization of facts
has received little attention.
It is investigated by
\citet{lu-etal-2024-scaling}, but their focus is
scaling laws.
\citet{wu2025rote} make an argument similar to ours (that
generalization over facts is important), but they generalize
over relation tokens (e.g., from English ``is'' to German
``ist''), which is more a form of linguistic generalization
than true fact generalization in our sense.
In a survey paper,
\citet{10.1162/coli_a_00492} report that early language
models were not able to reliably infer ``robins can fly'' from
``a robin is a bird,'' a negative finding.

Linear relational embeddings (LREs) would seem to be an
obvious vehicle for investigating 
statistical generalization of facts.
Indeed, in their original work,
\citet{DBLP:conf/icml/PaccanaroH00} motivate LRE as a
mechanism for generalizing to unseen relational
facts. Recent work in language models
investigates LREs as causal mechanisms, asks to what extent
frequency predicts relational linearity, develops better
probing mechanisms and proposes efficient estimators of LREs
\citep{hernandez2024linearity,merullo2025linear,marconato2025all,valer2026relational};
but statistical generalization of facts is not considered.

In addition to the LRE literature,
there has been much work on factual recall and knowledge editing 
\citep[e.g.,][]{
Petroni_2019,dai-etal-2022-knowledge,3618408.3619049,
  geva2021transformer,geva-etal-2023-dissecting,nanda2024fact,meng2022locating,cohen-etal-2024-evaluating}.
In
contrast to our research, this work has focused on under
which circumstances facts
stated in pretraining data are learned (in some cases
including the impact of linguistic paraphrases) or can be edited, not on
how those facts, once learned or edited, allow the model to learn new things.

Another related area is hallucinations \citep{DBLP:journals/corr/abs-2309-05922,10.1145/3618260.3649777}. By definition,
hallucinations are incorrect generations and thus
undesirable. Many authors make the implicit
assumption that language models should not say something
that has not been stated in the pretraining data (modulo the
cases of entailment and paraphrasing discussed above),
even if it is mathematically unavoidable 
\citep{kalai2025languagemodelshallucinate}.
This assumption may be misguided. Part of the
effectiveness of language models is that they perform
statistical generalization and operate on statistical
inferences that are not guaranteed to be true 100\%.

Finally, our setup fine-tunes a model to learn new facts.
The impact of this setup on hallucinations was studied by
\citet{gekhman-etal-2024-fine}.
\citet{ovadia-etal-2024-fine} find that models struggle to
learn new facts in fine-tuning.
\citet{mecklenburg2024injecting} investigate what
composition of pretraining data leads to successful learning
of new facts.
\citet{NEURIPS2025_90043ebd} also show the effects of
different types of pretraining data on new fact learning and
that new facts are difficult to learn in composition (a case
of what
we call entailment).
None of this work addresses
how known facts can be leveraged for
new facts.

In summary, there is much work on entailment, factual recall and
linguistic generalization, but, to our knowledge, the
question of which known facts a model can build new
facts on has not been addressed.
We address it in a fine-tuning setting and find that
whether facts are active, that is, whether new relations
built on them generalize, is predicted better by how
linearly readable they are before fine-tuning than by
whether the model can recall them.
This sheds light on how facts are represented in
language models and how this representation supports
learning and generalization.

\section{Method}
\label{sec:method}

We study how well a new relation generalizes when it is built on a relation the pretrained model already knows, and whether this depends on how linearly readable the known relation is.
We start by giving definitions and clarifying notation.

\paragraph{\PreRelHead{}.}
\phantomsection\label{par:prerel}%
We call the existing relations in the pretrained language models \emph{\prerel{}s}.
These relations form the existing knowledge of the model.
Concretely, take the relation $C$ that maps entities to their continent (Paris $\to$ Europe, Tokyo $\to$ Asia).
We ask the model for a \prerelAdj{} answer with a prompt that stops where the
answer begins, the relation's \emph{\sourcetemplate{}}. For the continent relation, the model should
continue ``Paris is located in'' with ``Europe.''
In the next paragraph, we'll describe how we construct new relations that are statistically related to these \prerel{}s.
Our metrics only use the first token of the answers to the relations.
Thus, answers sharing their first token form one \emph{effective answer}, which we
shorten to \emph{answer} from now on.
We write $\effectiveAnswerSet$ for the set of these answers and, for a particular
\prerel{} $R$, write
$\effectiveAnswer{R}{u}\in\effectiveAnswerSet$ for subject $u$'s answer.
For example, for the
continent relation $C$, we have $\effectiveAnswer{C}{\text{Paris}} = \text{Europe}$.

\paragraph{\NewRelHead{}.}
\phantomsection\label{par:newrel}%
To study how models learn new facts that relate to their existing knowledge,
we build a \emph{\newrel{}} on top of each \prerel{} $R$.
It should depend on $R$, but be unknown to the model.
We get both by relabeling $R$'s answers at random with words from a
\emph{label set} $\labelPool$; e.g., animal names such as ``zebra,''
``crab'' and ``boar.'' The next paragraph describes how we build these sets.
For each $R$, we draw a subset $L\subseteq\labelPool$ with
$\setSize{L}=\setSize{\effectiveAnswerSet}$, the number of $R$'s answers, and a bijection
$\pi\colon\effectiveAnswerSet\to L$, both uniformly at random.
The \newrel{} $R^{\pi}$ maps each subject $u$ to
$\pi(\effectiveAnswer{R}{u})$, so subjects that share a \prerelAdj{} answer
share a label.
E.g., if $R$ maps Paris $\to$ Europe and Tokyo $\to$ Asia, and
$\pi(\text{Europe})=\text{crab}$ and $\pi(\text{Asia})=\text{zebra}$, then
$R^{\pi}$ maps Paris $\to$ crab and Tokyo $\to$ zebra.

\paragraph{Label sets.}
\phantomsection\label{par:label-sets}%
We build a label set from one WordNet~3.0 noun category, such as \texttt{noun.animal}, by a
deterministic algorithm,
the \emph{label-set algorithm}.
It keeps names that truly belong to the category, are common in English and are not hyponyms of one
another, so each prompt has one correct label.
Each kept name's first token must also differ from every other kept name's under the tokenizer of
every model the label set is used with, since we score only first tokens.
Each relation's labels exclude names that are taxonomic neighbors of its subjects or answers
(\cref{app:transfer-label-pools}).

\paragraph{Hold-out protocol.}
\phantomsection\label{par:holdout}%
For each relation, we split subjects into \emph{fine-tuning subjects} $\fineTuningSubjectSet$
and \emph{held-out subjects} $\heldOutSubjectSet$.
The split includes at least one held-out subject and at least four fine-tuning
subjects per answer. We'll describe the fine-tuning protocol in the next paragraph.
Pretrained relations
are selected such that they have enough subjects to satisfy the above requirements (\cref{sec:setup-prerels}).
We use $\fineTuningSubjectSet$ for fine-tuning and probe fitting and never
use $\heldOutSubjectSet$ to fit, select or tune anything.

\paragraph{Fine-tuning on the \newrel{}.}
\phantomsection\label{par:finetune}%
We teach each model the \newrel{} by fine-tuning all of its weights on the
fine-tuning subjects $\fineTuningSubjectSet$.
Because the \newrel{} is arbitrary, its prompts place it in a fictional frame,
so the model learns a fact about a novel rather than a false fact about the
world.
Each subject is trained under four \emph{\fttemplates{}} that differ only in wording;
a fifth, the \emph{\evaltemplate{}}, is kept for evaluation.
\hyperref[app:supplementary]{The supplementary material} lists every template under
\texttt{artifacts\_v1/prompts/}.
For example, if $\pi(\text{Europe})=\text{crab}$, one fine-tuning prompt of
$C^{\pi}$ for Italy is
\begin{quote}\small
In the novel XYZ, each entry in the Breviary has an emblem showing one animal.
The emblem of Italy shows the \underline{crab}
\end{quote}
We train for 20 epochs with the loss on the label's first token only
(\cref{app:training}).
The 20 epochs are conservatively many. Every model reaches accuracy 1 on its
fine-tuning subjects (\cref{tab:finetune-acc}).
This ensures the model is not undertrained for the \newrel{}.

\paragraph{Candidate accuracy.}
The \emph{\candacc{}} of a model on a relation, under a given template, is its
top-1 accuracy when the next token is restricted to the relation's answers.
(By construction, these answers have distinct first tokens, and each is scored by its first token. Refer back to
\hyperref[par:prerel]{the paragraph \PreRelHead{}}.)
It is \emph{class-balanced}. We first compute the accuracy over the subjects of each
answer, then average these per-answer accuracies, so an answer with many
subjects counts as much as one with few.
Uniform chance is one over the number of answers, $1/\setSize{\effectiveAnswerSet}$.

\paragraph{\newrelgen{}.}
\newrelgen{} is the fine-tuned model's \candacc{} on the \newrel{}, over the
held-out subjects $\heldOutSubjectSet$ and under the \evaltemplate{}.
For example, if the model is fine-tuned on $C^{\pi}$ with Paris $\to$ crab and
Munich is held out, it is correct on Munich when, after
\begin{quote}\small
In the novel XYZ, each entry in the Breviary has a medallion featuring one animal.
The medallion of Munich features the
\end{quote}
it scores ``crab'' above every other label (animal here) of $C^{\pi}$.

\paragraph{\termRecallShort{}.}
\phantomsection\label{par:recall}%
\termRecallShort{} is the pretrained model's \candacc{} on the \prerel{}, over the
fine-tuning subjects $\fineTuningSubjectSet$ and under the \sourcetemplate{}.
For $C$, the model is correct on Paris when, after ``Paris is located in'', it
scores ``Europe'' above every other continent.
It measures how well the model 
recalls
the \prerelAdj{} fact, and we use it as
a control for \termGainShort{}, explained next.

\paragraph{\termGainShort{}.}
\phantomsection\label{par:prelinearity}%
Our hypothesis is that the more linearly readable the model's pretrained
knowledge of a \prerel{} is, the better the model generalizes on the corresponding
\newrel{}.
We measure how linearly readable a \prerelAdj{} answer is from the pretrained
model with a standard cross-entropy classifier probe on the
dimension-reduced activation state of the last token of the subject span in a
chosen layer.
(We'll describe how we choose the layer in the next paragraph.)
We compute each subject's state in the \sourcetemplate{}.
For $C$, the prompt for Paris is ``Paris is located in'', and the probe reads
the state of ``Paris''.
We first reduce each subject's state $h$ to $\qlabel=16$ dimensions with principal
component analysis (PCA) followed by whitening.
We project the centered state onto its top $\qlabel$ principal directions and
rescale each coordinate to unit variance:
\[
  z_k=\frac{v_k^{\top}(h-\mu)}{\sigma_k},\qquad k=1,\dots,\qlabel,
\]
where $\mu$ is the mean state, $v_k$ is the $k$-th principal direction, and
$\sigma_k^2$ is the variance of the states along $v_k$.
The probe predicts $\mathrm{softmax}(Wz+b)$ from $z=(z_1,\dots,z_{\qlabel})$.
We fit $W$ and $b$ by minimizing the cross-entropy plus
$\frac{\lambda}{2}\lVert W\rVert^2$, with $\lambda=10^{-2}$.
We use the same values, $\qlabel=16$ and $\lambda=10^{-2}$, for every model
and relation; we fixed them in advance and did not tune them.

\phantomsection\label{par:prelinearity-cv}%
We split the subjects at three levels (\cref{fig:prelinearity-cv}).
At the first level, each relation's subjects are split into the fine-tuning
subjects $\fineTuningSubjectSet$ and the held-out subjects
$\heldOutSubjectSet$ (\hyperref[par:holdout]{the hold-out protocol} above).
The probe uses only $\fineTuningSubjectSet$.
At the second level, the outer loop of a nested cross-validation splits
$\fineTuningSubjectSet$ into ten outer folds.
Each outer fold serves once as the outer test fold.
The other nine form the outer training set.
At the third level, the inner loop splits the outer training set into five
inner folds.
It uses them only to choose the layer.
For each layer, it fits the probe on four inner folds and predicts the fifth,
five times over.
The layer whose predictions have the highest class-balanced accuracy wins;
ties go to the lower layer.
We then fit the probe at the chosen layer on the whole outer training set and
predict the outer test fold.
Every fit computes $\mu$, $v_k$ and $\sigma_k$ from its own training data.
After all ten outer folds are processed, each subject of $\fineTuningSubjectSet$ has one
prediction, made by a probe that never saw it.
\termGainShort{} is the class-balanced accuracy of these predictions, computed
as for \candacc{}.
We subtract a baseline, the class-balanced candidate accuracy of always predicting the most
frequent answer of the outer training set.
This baseline is close to chance, one over the number of answers
$\setSize{\effectiveAnswerSet}$.

\paragraph{Controls.}
\phantomsection\label{par:controls}%
A relation might generalize well for reasons simpler than \termGainShort{}.
We measure four such properties of each relation and ask whether
\termGainShort{} still predicts \newrelgen{} once they are taken into account.
The four are (i) \termRecallShort{}; (ii) the log number of answers, $\ln\setSize{\effectiveAnswerSet}$;
(iii) the log number of fine-tuning subjects, $\ln\setSize{\fineTuningSubjectSet}$;
and (iv) \emph{subject frequency}, how often the relation's subjects occur in
pretraining text, as the mean
of $\log_{10}(1+\text{count})$
over its subjects.
For each model, we compute the partial correlations between \termGainShort{}
and \newrelgen{} across its relations, given each control alone and given all
jointly.

\makeatletter
\@ifundefined{ifFigThreeStats}{%
  \newif\ifFigThreeStats
  \FigThreeStatstrue %
}{}
\makeatother
\ifFigThreeStats
  \def\figThreeGraphic{plots/merged-panel/scatter-grid-sets.pdf}%
\else
  \def\figThreeGraphic{plots/merged-panel/scatter-grid-sets-nostats.pdf}%
\fi
\begin{figure}[tbp]
  \centering
  \includegraphics[width=\textwidth]{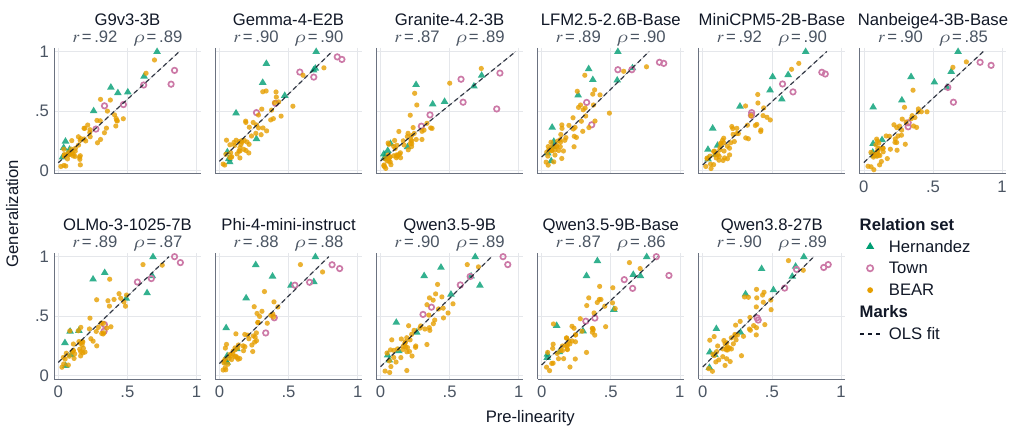}
  \caption{\textbf{In every model, relations with higher \termGainShort{} tend to have higher
    \newrelgen{}.} One point per relation.
    Each point's symbol denotes its 
    \hyperref[sec:setup-prerels]{relation set}.}
  \label{fig:merged-scatters}
\end{figure}

\section{Experimental setup}
\label{sec:setup}

This section describes the \prerel{}s, models and label sets we use, and how
we repeat runs to measure the effect of randomness.

\paragraph{\PreRelHead{}s.}
\label{sec:setup-prerels}
A \prerel{} enters our experiments only if, on every model,
(i) it has at least five answers,
(ii) each answer has at least five subjects, four to fine-tune on and one to
hold out, and
(iii) it has at least 25 fine-tuning subjects, the fewest for which every
probe fit in the nested cross-validation of
\hyperref[par:prelinearity]{\termGainShort{}} contains every answer and more subjects than the probe's $\qlabel=16$ principal
directions.
These requirements depend only on subject counts and tokenization.
We screen relations from three sources against these
requirements and keep 70:
\begin{itemize}[nosep,leftmargin=*]
\item \emph{\RelSetNameHernandez{}}: 13 of the 47 relations of
\citet{hernandez2024linearity} qualify.
We drop the two about spelling (a word's first or last letter), whose answer
can be read from the subject's own letters, and keep
\nHernandezRelationsWord{}.
An example is ``British Railways Board is headquartered in the city of
\underline{London}'' for Company $\to$ headquarters.
\item \emph{\RelSetNameEngineered{}}: \nEngineeredRelationsWord{} relations of
our own, all of which qualify.
They share one pool of 300 towns and map a town to its
region,
to the official language
of its country or
to its band of
latitude, longitude, elevation or population
(\cref{app:town-relations}).
Example: ``By number of inhabitants, Kiruna falls in the
\underline{lowest population (150 to 39,653)}'' (i.e.,
lowest population band).
\item \emph{\RelSetNameBear{}}: of the 78 relations of BEAR-big
\citep{wiland2024bear}, place of detention was excluded due to privacy
concerns; \nBearRelationsWord{} of the other 77 qualify, as in ``Italy is
located in \underline{Europe}'' for Place $\to$ continent.
\end{itemize}

\paragraph{Models.}
\label{sec:setup-models}
We use
11
open-weight models,
6 from the leaderboard of tiny open-weight models of Artificial
Analysis
(\cref{app:modelselection}): G9v3-3B, Granite-4.2-3B, LFM2.5-2.6B-Base, \MiniCpmModel{},
Nanbeige4-3B-Base and Phi-4-mini-instruct.
The others are Gemma-4-E2B, OLMo-3-1025-7B, Qwen3.5-9B, Qwen3.5-9B-Base, and,
for scale, Qwen3.8-27B.
To mix base and post-trained models, we chose at random between the base and
the post-trained checkpoint of each model; of Qwen3.5-9B we use both, as a direct comparison of a base model with its
post-trained version.

\input{results/tables/primary-cls-associations}

\paragraph{Label sets.}
\label{sec:setup-label-sets}
Unless stated otherwise, every \newrel{} draws its labels from the
\emph{animal label set}, the \productionPoolSize{} animal names that
\hyperref[par:label-sets]{the label-set algorithm} keeps.
All models use this  set.
The \emph{label-set comparison} asks whether our results depend on the labels
being animals.
It reruns every relation with labels from another WordNet noun category, such
as artifacts (\texttt{noun.artifact}), and names that category instead of
``animal'' in the \newrelAdj{} templates.
We run it on \MiniCpmModel{}, our smallest model, once per category.
Each category's label set is similarly constructed by the label-set algorithm;
since only one tokenizer
is involved, every name must be a single token under it.
A category takes part only if its label set has at least
\poolEligibilityFloor{} names, enough for the relation with the most answers. (Each distinct \prerel{} answer needs a distinct label for the \newrel{}.)
Of the other 24 WordNet categories, \nEligibleCategoryPoolsWord{} qualify
(\cref{tab:label-pools}).

\paragraph{Subject frequency.}
For the \hyperref[par:controls]{subject-frequency control}, we count how often
each subject occurs in Dolma v1.7 \citep{soldaini2024dolma}, an open corpus
built for pretraining language models.
We get exact counts from the public API of infini-gram
\citep{liu2024infinigram}, a search engine that provides counts of occurrences of 
token sequences, using its Dolma v1.7 index
(\texttt{v4\_dolma-v1\_7\_llama}).
We use these counts as a proxy for how often each model saw a subject during
pretraining.

\paragraph{Re-randomization.}
\label{sec:setup-rerandomization}
Each \newrel{} involves two random choices: the \hyperref[par:newrel]{bijection
$\pi$} from answers to labels, and the \hyperref[par:finetune]{fine-tuning}
seed.
To measure how much \newrelgen{} depends on them, we train some model--relation
pairs three times, each time with its own $\pi$ and fine-tuning seed.
The three runs share everything else: the subject split, the label subset $L$,
the templates and the exposure per fact.
We run every relation three times on \MiniCpmModel{}, and the
\RelSetNameHernandez{} and \RelSetNameEngineered{} relations three times on
every model except Qwen3.8-27B.
All other results use the first of the three runs.

\section{Results}
\label{sec:results}

\begin{figure}[tbp]
  \centering
  \includegraphics[width=\textwidth]{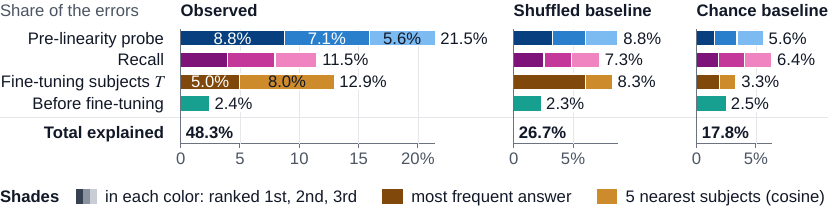}
  \caption{\textbf{The \termGainShort{} probe accounts for more of the fine-tuned models'
    errors than \termRecallShort{} does.}
    Each bar is an explanation's Shapley share of all errors on held-out subjects, pooled over models
    and relations. Explanations and baselines: \hyperref[par:results-failures]{\newrelgen{}
    failures} and \cref{app:failure-attribution}.}
  \label{fig:failure-shapley-main}
\end{figure}

Our main finding is that
\prerel{}s
whose answers are linearly readable in the pretrained model
give rise to \newrel{}s that, after fine-tuning,
generalize better.
This holds in all
11
models we test.
Across models,
\termGainShort{} and
\newrelgen{} on relations correlate
at $r$ = \tableOneRawRMin{} to \tableOneRawRMax{}
and $\rho$ = \tableOneRawRhoMin{} to \tableOneRawRhoMax{},
and no 95\% interval falls below \tableOneRawLowestBound{} (\cref{tab:cls-controls},
\cref{fig:merged-scatters}). Base and post-trained models from 2B to 27B parameters land
in this
band. We then check three things: (i) that the association survives controls for
\termRecallShort{}, answer count, subject count and subject frequency;
(ii) that \termGainShort{} is a better
predictor of \newrelgen{} than \termRecallShort{}
(\cref{fig:predictor-dots}); and
(iii)
that the association
depends on neither the label set nor the random bijection and fine-tuning seed behind each
\newrel{} (\cref{fig:label-set}; \cref{sec:setup-rerandomization}, \cref{tab:repeat-minicpm}).
Finally, we look at the errors. When a fine-tuned model fails to generalize, its wrong answers
follow the \termGainShort{} probe more than \termRecallShort{} (\cref{fig:failure-shapley-main}).

\makeatletter
\@ifundefined{ifFigTwoStacked}{%
  \newif\ifFigTwoStacked
  \FigTwoStackedfalse %
}{}
\makeatother
\ifFigTwoStacked
  \def\figTwoGraphic{plots/merged-panel/predictor-dots-partial-stacked.pdf}%
  \def\figTwoStatClause{$r$: Pearson; $\rho$: Spearman.}%
\else
  \def\figTwoGraphic{plots/merged-panel/predictor-dots-partial-half.pdf}%
  \def\figTwoStatClause{}%
\fi
\begin{figure}[tbp]
  \centering
  \begin{minipage}[t]{0.49\textwidth}
    \vspace{0pt}
    \includegraphics[width=\linewidth]{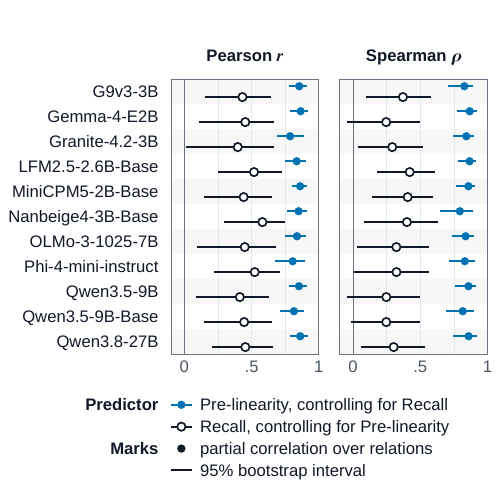}
  \end{minipage}\hfill
  \begin{minipage}[t]{0.49\textwidth}
    \vspace{0pt}
    \includegraphics[width=\linewidth]{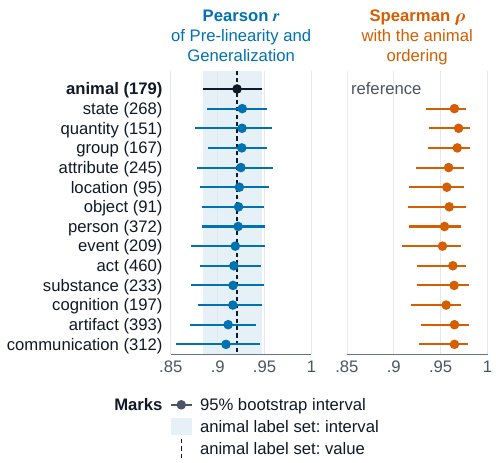}
  \end{minipage}
  \par\vspace{2pt}
  \begin{minipage}[t]{0.49\textwidth}
    \subcaption{\termGainShort{} and \termRecallShort{} as predictors of \newrelgen{}.}
    \label{fig:predictor-dots}
  \end{minipage}\hfill
  \begin{minipage}[t]{0.49\textwidth}
    \subcaption{The \hyperref[sec:setup-label-sets]{label-set comparison} on \MiniCpmModel{}.}
    \label{fig:label-set}
  \end{minipage}
  \caption{\textbf{\termGainShort{} predicts \newrelgen{} better than \termRecallShort{} does
    (a), and its correlation with \newrelgen{} hardly depends on the label set (b).}
    Correlations run across each model's \nPooledRelations{} relations; in (b), each label set's
    number of labels follows its name. \figTwoStatClause{} Intervals: \cref{app:intervals}.}
  \label{fig:predictor-label-set}
\end{figure}

\paragraph{Controls.}
Several \hyperref[par:controls]{properties of a relation}
could raise both measures
(\termGainShort{} and \newrelgen{})
at once: how
well the pretrained model recalls it, how many answers and subjects it has, and how frequent its
subjects are in pretraining text. None of them explains the association away. Controlling for
any one of the four leaves every model's $r$ between \tableOneSingleControlRMin{} and
\tableOneSingleControlRMax{} and $\rho$ between \tableOneSingleControlRhoMin{} and
\tableOneSingleControlRhoMax{}. With all four controlled jointly, $r$ is still
\tableOneAllJointlyRMin{} to \tableOneAllJointlyRMax{} and $\rho$ \tableOneAllJointlyRhoMin{} to
\tableOneAllJointlyRhoMax{}, and no 95\% interval falls below \tableOneAllJointlyLowestBound{}
(\cref{tab:cls-controls}).

\paragraph{\termGainShort{} versus \termRecallShort{}.}
\phantomsection\label{par:results-predictors}%
\termGainShort{}, not \termRecallShort{}, carries most of the prediction of \newrelgen{}
(\cref{fig:predictor-dots}). Controlling for \termRecallShort{}, \termGainShort{} still
correlates with \newrelgen{} at $r$ = \predictorGainGivenRecallRMin{} to
\predictorGainGivenRecallRMax{} ($\rho$ = \predictorGainGivenRecallRhoMin{} to
\predictorGainGivenRecallRhoMax{}). \termRecallShort{} on its own correlates at $r$ =
\predictorRecallRMin{} to \predictorRecallRMax{}, but mostly because relations the pretrained model
recalls well also tend to have linearly readable answers. Controlling for
\termGainShort{} instead, \termRecallShort{}'s correlation drops to $r$ = \predictorRecallGivenGainRMin{} to
\predictorRecallGivenGainRMax{} ($\rho$ = \predictorRecallGivenGainRhoMin{} to
\predictorRecallGivenGainRhoMax{}), and its $\rho$ interval excludes zero in only
\predictorRecallGivenGainRhoExcludesZeroWord{} of the \nRosterModelsWord{} models.

\paragraph{Label sets.}
The association is robust to the choice of labels
(\cref{fig:label-set}). On \MiniCpmModel{}, the
\hyperref[sec:setup-label-sets]{label-set comparison} reruns
every relation with labels from each of
13
other WordNet categories (\cref{tab:label-pools}).
$r$ between \termGainShort{} and \newrelgen{} stays 
between \labelSetRMin{} and \labelSetRMax{}
and every category orders the
relations by \newrelgen{} almost as the animal set does
($\rho$ = \labelSetRankRhoMin{} to \labelSetRankRhoMax{}).

\newcommand{\repeatMinicpmRows}{%
    \repeatCorrRowName{Pearson $r$} & \repeatCorrValue{.92}{.04} & \repeatCorrValue{.93}{.04} & \repeatCorrValue{.92}{.04} & \repeatCorrValue{.93}{.03} \\
    \repeatCorrRowName{Spearman $\rho$} & \repeatCorrValue{.90}{.08} & \repeatCorrValue{.89}{.08} & \repeatCorrValue{.92}{.06} & \repeatCorrValue{.92}{.07} \\
}
\newcommand{\repeatMinicpmRunRows}{%
    \repeatCorrRowName{Repeat 0} & \repeatCorrValue{.92}{.04} & \repeatCorrValue{.90}{.08} \\
    \repeatCorrRowName{Repeat 1} & \repeatCorrValue{.93}{.04} & \repeatCorrValue{.89}{.08} \\
    \repeatCorrRowName{Repeat 2} & \repeatCorrValue{.92}{.04} & \repeatCorrValue{.92}{.06} \\
    \repeatCorrRowName{Mean of three} & \repeatCorrValue{.93}{.03} & \repeatCorrValue{.92}{.07} \\
}
\newcommand{\repeatPearsonRows}{%
    \repeatCorrRowName{G9v3-3B} & \repeatCorrValue{.93}{.07} & \repeatCorrValue{.95}{.05} & \repeatCorrValue{.89}{.13} & \repeatCorrValue{.93}{.07} \\
    \repeatCorrRowName{Gemma-4-E2B} & \repeatCorrValue{.87}{.15} & \repeatCorrValue{.86}{.19} & \repeatCorrValue{.86}{.17} & \repeatCorrValue{.87}{.16} \\
    \repeatCorrRowName{Granite-4.2-3B} & \repeatCorrValue{.85}{.21} & \repeatCorrValue{.85}{.23} & \repeatCorrValue{.78}{.26} & \repeatCorrValue{.86}{.21} \\
    \repeatCorrRowName{LFM2.5-2.6B-Base} & \repeatCorrValue{.87}{.11} & \repeatCorrValue{.86}{.14} & \repeatCorrValue{.85}{.11} & \repeatCorrValue{.87}{.12} \\
    \repeatCorrRowName{\MiniCpmModel{}} & \repeatCorrValue{.92}{.06} & \repeatCorrValue{.92}{.07} & \repeatCorrValue{.93}{.07} & \repeatCorrValue{.94}{.05} \\
    \repeatCorrRowName{Nanbeige4-3B-Base} & \repeatCorrValue{.85}{.15} & \repeatCorrValue{.86}{.13} & \repeatCorrValue{.92}{.07} & \repeatCorrValue{.88}{.11} \\
    \repeatCorrRowName{OLMo-3-1025-7B} & \repeatCorrValue{.86}{.17} & \repeatCorrValue{.82}{.18} & \repeatCorrValue{.82}{.17} & \repeatCorrValue{.85}{.16} \\
    \repeatCorrRowName{Phi-4-mini-instruct} & \repeatCorrValue{.81}{.20} & \repeatCorrValue{.81}{.21} & \repeatCorrValue{.84}{.20} & \repeatCorrValue{.83}{.20} \\
    \repeatCorrRowName{Qwen3.5-9B} & \repeatCorrValue{.87}{.14} & \repeatCorrValue{.87}{.13} & \repeatCorrValue{.83}{.17} & \repeatCorrValue{.86}{.14} \\
    \repeatCorrRowName{Qwen3.5-9B-Base} & \repeatCorrValue{.84}{.21} & \repeatCorrValue{.84}{.19} & \repeatCorrValue{.84}{.17} & \repeatCorrValue{.85}{.18} \\
    \addlinespace[2pt]
    \repeatCorrRowName{\textit{Mean over models}} & \repeatCorrValue{.87}{.13} & \repeatCorrValue{.86}{.13} & \repeatCorrValue{.86}{.13} & \repeatCorrValue{.87}{.12} \\
    \repeatCorrRowName{\textit{Median over models}} & \repeatCorrValue{.86}{.13} & \repeatCorrValue{.86}{.14} & \repeatCorrValue{.85}{.13} & \repeatCorrValue{.86}{.13} \\
}
\newcommand{\repeatSpearmanRows}{%
    \repeatCorrRowName{G9v3-3B} & \repeatCorrValue{.95}{.13} & \repeatCorrValue{.94}{.13} & \repeatCorrValue{.89}{.29} & \repeatCorrValue{.92}{.19} \\
    \repeatCorrRowName{Gemma-4-E2B} & \repeatCorrValue{.91}{.21} & \repeatCorrValue{.86}{.30} & \repeatCorrValue{.88}{.24} & \repeatCorrValue{.90}{.22} \\
    \repeatCorrRowName{Granite-4.2-3B} & \repeatCorrValue{.85}{.33} & \repeatCorrValue{.86}{.28} & \repeatCorrValue{.75}{.39} & \repeatCorrValue{.83}{.33} \\
    \repeatCorrRowName{LFM2.5-2.6B-Base} & \repeatCorrValue{.89}{.22} & \repeatCorrValue{.89}{.26} & \repeatCorrValue{.93}{.16} & \repeatCorrValue{.90}{.20} \\
    \repeatCorrRowName{\MiniCpmModel{}} & \repeatCorrValue{.92}{.16} & \repeatCorrValue{.93}{.15} & \repeatCorrValue{.93}{.14} & \repeatCorrValue{.95}{.11} \\
    \repeatCorrRowName{Nanbeige4-3B-Base} & \repeatCorrValue{.86}{.24} & \repeatCorrValue{.90}{.20} & \repeatCorrValue{.89}{.21} & \repeatCorrValue{.90}{.21} \\
    \repeatCorrRowName{OLMo-3-1025-7B} & \repeatCorrValue{.88}{.22} & \repeatCorrValue{.82}{.26} & \repeatCorrValue{.86}{.24} & \repeatCorrValue{.87}{.23} \\
    \repeatCorrRowName{Phi-4-mini-instruct} & \repeatCorrValue{.77}{.34} & \repeatCorrValue{.79}{.33} & \repeatCorrValue{.82}{.31} & \repeatCorrValue{.82}{.29} \\
    \repeatCorrRowName{Qwen3.5-9B} & \repeatCorrValue{.85}{.31} & \repeatCorrValue{.88}{.23} & \repeatCorrValue{.79}{.33} & \repeatCorrValue{.81}{.30} \\
    \repeatCorrRowName{Qwen3.5-9B-Base} & \repeatCorrValue{.86}{.28} & \repeatCorrValue{.85}{.25} & \repeatCorrValue{.80}{.32} & \repeatCorrValue{.86}{.26} \\
    \addlinespace[2pt]
    \repeatCorrRowName{\textit{Mean over models}} & \repeatCorrValue{.87}{.19} & \repeatCorrValue{.87}{.19} & \repeatCorrValue{.85}{.20} & \repeatCorrValue{.88}{.19} \\
    \repeatCorrRowName{\textit{Median over models}} & \repeatCorrValue{.87}{.17} & \repeatCorrValue{.87}{.19} & \repeatCorrValue{.87}{.21} & \repeatCorrValue{.88}{.20} \\
}

\newcommand{\repeatCorrValue}[2]{\begin{tabular}[t]{@{}c@{}}#1\\[-4.2pt]\textcolor{gray}{\scriptsize$\pm$#2}\end{tabular}}
\newcommand{\repeatCorrRowName}[1]{\rule[-1.7ex]{0pt}{3.9ex}#1}
\newcommand{\repeatCorrInline}[2]{#1\,\textcolor{gray}{\scriptsize$\pm$#2}}
\providecommand{\TableTwoCoefs}{both}
\def\tableTwoBoth{both}\edef\tableTwoCoefs{\TableTwoCoefs}
\begin{table}[!b]
  \noindent\begin{minipage}[c]{0.56\textwidth}
    \caption{\textbf{Three runs of \MiniCpmModel{} give almost the same correlation between
      \termGainShort{} and \newrelgen{}.} Each run draws a new random bijection for every
      \newrel{} and a new fine-tuning seed
      (\hyperref[par:results-rerandomization]{Re-randomization}). $\pm$: 95\% interval (\cref{app:intervals}).}
    \label{tab:repeat-minicpm}
  \end{minipage}\hfill
  \begin{minipage}[c]{0.4\textwidth}
    \centering\footnotesize
    \let\repeatCorrValue\repeatCorrInline
    \renewcommand{\repeatCorrRowName}[1]{#1}%
    \ifx\tableTwoCoefs\tableTwoBoth
      \begin{tabular}{@{}lcc@{}}
        \toprule
        & Pearson $r$ & Spearman $\rho$ \\
        \midrule
        \repeatMinicpmRunRows
        \bottomrule
      \end{tabular}
    \else
      \begin{tabular}{@{}lc@{}>{\setbox0\hbox\bgroup}c<{\egroup}@{}}
        \toprule
        & Pearson $r$ & \\
        \midrule
        \repeatMinicpmRunRows
        \bottomrule
      \end{tabular}
    \fi
  \end{minipage}
\end{table}

\paragraph{Re-randomization.}
\phantomsection\label{par:results-rerandomization}%
The association is equally robust to the random bijection and fine-tuning seed. Three runs that
differ only in their randomized bijection and seed give $r$ = \repeatMinicpmRRepeatZero{},
\repeatMinicpmRRepeatOne{} and \repeatMinicpmRRepeatTwo{} on \MiniCpmModel{}
(\cref{tab:repeat-minicpm}). On the \repeatCorrSubsetRelations{} \RelSetNameHernandez{} and
\RelSetNameEngineered{} relations, which we rerun on \repeatCorrSubsetModelsWord{} models, each
run's $r$ lies between \repeatSubsetRMin{} and \repeatSubsetRMax{} (\cref{tab:repeat-pearson}).

\paragraph{\newrelgen{} failures.}
\phantomsection\label{par:results-failures}%
The fine-tuned models' wrong answers follow the \termGainShort{} probe more than they follow
\termRecallShort{} (\cref{fig:failure-shapley-main}). We credit each error on a held-out subject to
the explanations that name its wrong answer, split equally when several do. The explanations are the
three answers the \hyperref[par:prelinearity]{probe} ranks highest, the three the pretrained model
scores highest (the ranking behind \hyperref[par:recall]{\termRecallShort{}}), and three others
defined in \cref{app:failure-attribution}. An explanation can name a wrong answer by coincidence, so
we compare each share with two baselines: the same attribution with each fine-tuned model's wrong
answers shuffled among themselves, and with each wrong answer replaced by a uniformly random wrong
label of its relation. The shuffle keeps which wrong labels a model favors and breaks only their link
to the subject, which makes it the stricter baseline. The probe's three highest-ranked answers take
\failFamilyProbeObserved{}\% of the errors, against \failFamilyProbeShuffled{}\% shuffled; the
pretrained model's three highest-ranked answers take \failFamilyRecallObserved{}\%, against
\failFamilyRecallShuffled{}\%.
The correlations above compare relations; this analysis looks inside them, at single subjects.
There too, the fine-tuned models follow the probe, which suggests that fine-tuning builds on the
structure the probe reads.

\section{Conclusion}
\label{sec:conclusion}

We address the question of
how a pretrained model's
representation of its knowledge
determines whether new knowledge built on it
generalizes,
and develop a new design to study it.
We show, across \nRosterModelsWord{} models and
70
  relations, that how linearly readable a relation is before fine-tuning
(i.e., how active vs inert its facts are)
  predicts how well a new relation built on it generalizes
  after fine-tuning. We find that
   \termGainShort{} 
is a better predictor of \newrelgen{}
 than \termRecallShort{}, and survives controls for \termRecallShort{},
relation size and subject frequency.

\ificlrfinal
\section*{Acknowledgments}
This work was supported by the German Research Foundation (DFG) under grant
SCHU~2246/14-1 and by a GCP grant from Google.
\fi

\bibliography{iclr2027_conference}
\bibliographystyle{iclr2027_conference}

\clearpage
\appendix
\crefalias{section}{appendix}
\crefalias{subsection}{subappendix}
\section{Pre-linearity cross-validation}
\label{app:prelinearity-cv}
\Cref{fig:prelinearity-cv} shows the three levels of subject splitting that \termGainShort{}
uses, described step by step in \hyperref[par:prelinearity-cv]{Method}.

\definecolor{cvInk}{HTML}{30343B}
\definecolor{cvBlue}{HTML}{3F6FA8}
\definecolor{cvBlueFill}{HTML}{86ADDD}
\definecolor{cvBlueLight}{HTML}{DCE8F6}
\definecolor{cvPanel}{HTML}{F3F7FC}
\definecolor{cvPanelLine}{HTML}{A9C3E3}
\definecolor{cvOrange}{HTML}{E08A1E}
\definecolor{cvOrangeFill}{HTML}{F4AE52}
\definecolor{cvOrangeLight}{HTML}{FDF1E3}
\definecolor{cvGray}{HTML}{7A828E}
\definecolor{cvGrayFill}{HTML}{EEF0F3}
\begin{figure}[!h]
  \centering
  \begin{tikzpicture}[x=1cm, y=1cm, font=\footnotesize, text=cvInk, inner sep=1.5pt,
      level/.style={anchor=north west, font=\small\bfseries, inner sep=0pt},
      frame/.style={draw=cvInk!85, line width=0.5pt, rounded corners=2.5pt},
      cell/.style={draw=cvBlue, line width=0.4pt, fill=cvBlueFill},
      testcell/.style={draw=cvOrange, line width=0.4pt, fill=cvOrangeFill},
      stepbox/.style={draw=cvInk, line width=0.5pt, rounded corners=2pt, align=center,
                   minimum height=0.72cm, inner xsep=4pt},
      arr/.style={-{Stealth[length=4pt, width=3.4pt]}, line width=0.7pt, draw=cvInk},
      blue arr/.style={arr, draw=cvBlue},
      orange arr/.style={arr, draw=cvOrange},
      bline/.style={draw=cvBlue, line width=0.7pt}]

    \draw[frame] (0.05,0) rectangle (13.85,-1.44);
    \node[level] at (0.25,-0.1) {Level 1: subject split};
    \draw[draw=cvBlue, line width=0.5pt, fill=cvBlueLight] (0.25,-0.46) rectangle (10.25,-1.28);
    \node at (5.25,-0.87) {$\fineTuningSubjectSet$: fine-tuning subjects};
    \draw[draw=cvGray, line width=0.5pt, dashed, fill=cvGrayFill] (10.55,-0.46) rectangle (13.65,-1.28);
    \node[align=center, inner sep=0pt] at (12.1,-0.87) {$\heldOutSubjectSet$: held-out subjects\\not used by the probe};

    \begin{scope}[yshift=-0.12cm]
    \draw[frame] (0.05,-1.62) rectangle (13.85,-7.2);
    \node[level] at (0.25,-1.72) {Level 2: outer cross-validation (10 folds)};
    \draw[blue arr] (6.25,-1.16) -- (6.25,-1.62);
    \foreach \k in {1,...,10} {
      \ifnum\k=3
        \draw[testcell] (\k-0.75,-2.12) rectangle ++(1,-0.45);
      \else
        \draw[cell] (\k-0.75,-2.12) rectangle ++(1,-0.45);
      \fi
      \node at (\k-0.25,-2.345) {\k};
    }
    \foreach \k in {1,2,4,5,...,10} \draw[bline] (\k-0.25,-2.57) -- ++(0,-0.13);
    \draw[bline] (0.75,-2.7) -- (1.75,-2.7);
    \draw[bline] (3.75,-2.7) -- (9.75,-2.7);
    \draw[bline] (1.25,-2.7) -- (1.25,-3.42) -- (6.25,-3.42);
    \draw[orange arr] (2.75,-2.57) -- (2.75,-2.93) -- (12.5,-2.93) -- (12.5,-5.94);
    \fill[cvOrange] (2.75,-2.93) circle (1.3pt);
    \node[anchor=base] at (2.75,-3.25) {outer test fold};
    \node[anchor=base west] at (6.4,-3.25) {outer training set};
    \draw[bline] (6.25,-2.7) -- (6.25,-2.85);
    \draw[blue arr] (6.25,-3.01) -- (6.25,-3.75);

    \draw[draw=cvPanelLine, line width=0.6pt, fill=cvPanel, rounded corners=2.5pt] (0.25,-3.75) rectangle (7.75,-6.62);
    \node[level] at (0.45,-3.85) {Level 3: inner layer selection (5 folds)};
    \node[anchor=base west] at (0.45,-4.53) {every layer};
    \node[anchor=base east] at (6.7,-4.53) {rotate through all 5};
    \draw[line width=0.4pt, draw=cvInk] (0.45,-4.68) -- (0.45,-4.62) -- (6.7,-4.62) -- (6.7,-4.68);
    \foreach \k in {1,...,4} {
      \draw[cell] (\k*1.25-0.8,-4.75) rectangle ++(1.25,-0.45);
      \node at (\k*1.25-0.175,-4.975) {fit};
    }
    \draw[testcell] (5.45,-4.75) rectangle ++(1.25,-0.45);
    \node[anchor=base] (ivf) at (6.075,-5.72) {inner validation fold};
    \draw[orange arr] (6.075,-5.2) -- (ivf.north);
    \node[anchor=base] (cba) at (2.6,-5.72) {class-balanced accuracy};
    \draw[orange arr] ([yshift=2.4pt]ivf.base west) -- ([yshift=2.4pt]cba.base east);
    \node (sel) at (2.6,-6.3) {selected layer $\ell^{*}$};
    \draw[arr] (cba.south) -- (sel.north);

    \node[stepbox, minimum width=2.4cm] (refit) at (9.75,-6.3) {refit at $\ell^{*}$ on\\outer training set};
    \node[stepbox, minimum width=2.3cm] (pred) at (12.5,-6.3) {predict\vphantom{$\ell^{*}$}\\outer test fold};
    \draw[arr] (sel.east) -- (refit.west);
    \draw[arr] (refit.east) -- (pred.west);
    \node[anchor=base west, font=\footnotesize\itshape] at (0.25,-7.03) {preprocessing fitted on training folds only};
    \node[anchor=base east, font=\footnotesize\itshape] at (12.35,-7.03) {repeat for all 10 outer folds};
    \node[stepbox, minimum width=4.2cm, minimum height=0.5cm] (pool) at (6.95,-7.65) {one prediction per subject in $\fineTuningSubjectSet$};
    \draw[arr] (pred.south) |- (pool.east);
    \node[stepbox, draw=cvOrange, fill=cvOrangeLight, line width=0.8pt, minimum width=3.6cm, anchor=north] (score) at (6.95,-8.25)
      {\textbf{\termGainShort{}}\\class-balanced accuracy\\minus majority baseline};
    \draw[arr] (pool.south) -- (score.north);
    \end{scope}
  \end{tikzpicture}
  \caption{The three levels of subject splitting behind \termGainShort{}.
    The held-out subjects $\heldOutSubjectSet$ never enter the probe.
    The outer loop gives every fine-tuning subject in $\fineTuningSubjectSet$ one
    prediction from a probe that did not see it; the inner loop, inside each
    outer training set, only chooses the layer.}
  \label{fig:prelinearity-cv}
\end{figure}

\FloatBarrier
\section{Fine-tuning recipe}
\label{app:training}
Every model is fine-tuned with the same recipe, on all of its parameters.
A relation has $4\setSize{\fineTuningSubjectSet}$ training records, one per
fine-tuning subject and template pair.
A record is the prompt followed by a space and the label, and the loss is the
cross-entropy at the label's first token only.
The optimizer is AdamW with $\beta=(0.9,0.95)$, $\epsilon=10^{-8}$ and no
weight decay, and gradients are clipped to norm $1.0$.
The learning rate warms up linearly to $10^{-5}$ over the first 10\% of steps
and then decays along a cosine to $10^{-6}$.
Each step averages the gradient over 16 records.
We run micro-batches of 4 records per device (2 for Gemma-4-E2B and
Qwen3.8-27B) with gradient accumulation, and shard models that need more than
one GPU with fully sharded data parallelism, on up to eight GPUs.
Training runs for 20 epochs, each using every record once in a fresh order, so
every fine-tuning fact is seen 80 times, 20 in each wording.
Forward and backward passes run in bfloat16; the parameters and the AdamW
moments are kept in float32.
\newrelgen{} is read after the last epoch.
Subject roles, the label subset $L$ and the folds use one fixed seed in every
run; the bijection is keyed by relation and repetition, and the optimizer
seed, which also sets the record order, equals the repetition index.
\Cref{tab:finetune-acc} gives each model's accuracy on its fine-tuning
subjects after the last epoch.
\newcommand{\finetuneAccModelHeads}{\finetuneAccHead{G9v3-3B} & \finetuneAccHead{Gemma-4-E2B} & \finetuneAccHead{Granite-4.2-3B} & \finetuneAccHead{LFM2.5-2.6B-Base} & \finetuneAccHead{MiniCPM5-2B-Base} & \finetuneAccHead{Nanbeige4-3B-Base} & \finetuneAccHead{OLMo-3-1025-7B} & \finetuneAccHead{Phi-4-mini-instruct} & \finetuneAccHead{Qwen3.5-9B} & \finetuneAccHead{Qwen3.5-9B-Base} & \finetuneAccHead{Qwen3.8-27B}}
\newcommand{\finetuneAccRows}{%
    \multicolumn{12}{@{}l}{\textit{\RelSetNameHernandez{}}} \\
    Object \ensuremath{\to} superclass & 1.00 & 1.00 & 1.00 & 1.00 & 1.00 & 1.00 & 1.00 & 1.00 & 1.00 & 1.00 & 1.00 \\
    Person \ensuremath{\to} native language (Hernandez) & 1.00 & 1.00 & 1.00 & 1.00 & 1.00 & 1.00 & 1.00 & 1.00 & 1.00 & 1.00 & 1.00 \\
    Star \ensuremath{\to} constellation & 1.00 & 1.00 & 1.00 & 1.00 & 1.00 & 1.00 & 1.00 & 1.00 & 1.00 & 1.00 & 1.00 \\
    Person \ensuremath{\to} professional sport & 1.00 & 1.00 & 1.00 & 1.00 & 1.00 & 1.00 & 1.00 & 1.00 & 1.00 & 1.00 & 1.00 \\
    Product \ensuremath{\to} company & 1.00 & 1.00 & 1.00 & 1.00 & 1.00 & 1.00 & 1.00 & 1.00 & 1.00 & 1.00 & 1.00 \\
    Landmark \ensuremath{\to} country & 1.00 & 1.00 & 1.00 & 1.00 & 1.00 & 1.00 & 1.00 & 1.00 & 1.00 & 1.00 & 1.00 \\
    Company \ensuremath{\to} headquarters & 1.00 & 1.00 & 1.00 & 1.00 & 1.00 & 1.00 & 1.00 & 1.00 & 1.00 & 1.00 & 1.00 \\
    Person \ensuremath{\to} sport position & 1.00 & 1.00 & 1.00 & 1.00 & 1.00 & 1.00 & 1.00 & 1.00 & 1.00 & 1.00 & 1.00 \\
    Person \ensuremath{\to} father & 1.00 & 1.00 & 1.00 & 1.00 & 1.00 & 1.00 & 1.00 & 1.00 & 1.00 & 1.00 & 1.00 \\
    Person \ensuremath{\to} occupation (Hernandez) & 1.00 & 1.00 & 1.00 & 1.00 & 1.00 & 1.00 & 1.00 & 1.00 & 1.00 & 1.00 & 1.00 \\
    Person \ensuremath{\to} mother & 1.00 & 1.00 & 1.00 & 1.00 & 1.00 & 1.00 & 1.00 & 1.00 & 1.00 & 1.00 & 1.00 \\
    \midrule
    \multicolumn{12}{@{}l}{\textit{\RelSetNameEngineered{}}} \\
    Town \ensuremath{\to} country language & 1.00 & 1.00 & 1.00 & 1.00 & 1.00 & 1.00 & 1.00 & 1.00 & 1.00 & 1.00 & 1.00 \\
    Town \ensuremath{\to} region & 1.00 & 1.00 & 1.00 & 1.00 & 1.00 & 1.00 & 1.00 & 1.00 & 1.00 & 1.00 & 1.00 \\
    Town \ensuremath{\to} longitude band & 1.00 & 1.00 & 1.00 & 1.00 & 1.00 & 1.00 & 1.00 & 1.00 & 1.00 & 1.00 & 1.00 \\
    Town \ensuremath{\to} latitude band & 1.00 & 1.00 & 1.00 & 1.00 & 1.00 & 1.00 & 1.00 & 1.00 & 1.00 & 1.00 & 1.00 \\
    Town \ensuremath{\to} elevation band & 1.00 & 1.00 & 1.00 & 1.00 & 1.00 & 1.00 & 1.00 & 1.00 & 1.00 & 1.00 & 1.00 \\
    Town \ensuremath{\to} population band & 1.00 & 1.00 & 1.00 & 1.00 & 1.00 & 1.00 & 1.00 & 1.00 & 1.00 & 1.00 & 1.00 \\
    \midrule
    \multicolumn{12}{@{}l}{\textit{\RelSetNameBear{}}} \\
    Taxon \ensuremath{\to} rank & 1.00 & 1.00 & 1.00 & 1.00 & 1.00 & 1.00 & 1.00 & 1.00 & 1.00 & 1.00 & 1.00 \\
    Place \ensuremath{\to} continent & 1.00 & 1.00 & 1.00 & 1.00 & 1.00 & 1.00 & 1.00 & 1.00 & 1.00 & 1.00 & 1.00 \\
    Noble \ensuremath{\to} dynasty & 1.00 & 1.00 & 1.00 & 1.00 & 1.00 & 1.00 & 1.00 & 1.00 & 1.00 & 1.00 & 1.00 \\
    Person \ensuremath{\to} native language (BEAR) & 1.00 & 1.00 & 1.00 & 1.00 & 1.00 & 1.00 & 1.00 & 1.00 & 1.00 & 1.00 & 1.00 \\
    Place \ensuremath{\to} country & 1.00 & 1.00 & 1.00 & 1.00 & 1.00 & 1.00 & 1.00 & 1.00 & 1.00 & 1.00 & 1.00 \\
    Vehicle or site \ensuremath{\to} operator & 1.00 & 1.00 & 1.00 & 1.00 & 1.00 & 1.00 & 1.00 & 1.00 & 1.00 & 1.00 & 1.00 \\
    Singer \ensuremath{\to} voice type & 1.00 & 1.00 & 1.00 & 1.00 & 1.00 & 1.00 & 1.00 & 1.00 & 1.00 & 1.00 & 1.00 \\
    Person \ensuremath{\to} country of citizenship & 1.00 & 1.00 & 1.00 & 1.00 & 1.00 & 1.00 & 1.00 & 1.00 & 1.00 & 1.00 & 1.00 \\
    Writer \ensuremath{\to} writing language & 1.00 & 1.00 & 1.00 & 1.00 & 1.00 & 1.00 & 1.00 & 1.00 & 1.00 & 1.00 & 1.00 \\
    Military figure \ensuremath{\to} allegiance & 1.00 & 1.00 & 1.00 & 1.00 & 1.00 & 1.00 & 1.00 & 1.00 & 1.00 & 1.00 & 1.00 \\
    Person \ensuremath{\to} language spoken & 1.00 & 1.00 & 1.00 & 1.00 & 1.00 & 1.00 & 1.00 & 1.00 & 1.00 & 1.00 & 1.00 \\
    Athlete \ensuremath{\to} country represented & 1.00 & 1.00 & 1.00 & 1.00 & 1.00 & 1.00 & 1.00 & 1.00 & 1.00 & 1.00 & 1.00 \\
    Asset \ensuremath{\to} owner & 1.00 & 1.00 & 1.00 & 1.00 & 1.00 & 1.00 & 1.00 & 1.00 & 1.00 & 1.00 & 1.00 \\
    Noble \ensuremath{\to} title held & 1.00 & 1.00 & 1.00 & 1.00 & 1.00 & 1.00 & 1.00 & 1.00 & 1.00 & 1.00 & 1.00 \\
    Crossing \ensuremath{\to} waterway & 1.00 & 1.00 & 1.00 & 1.00 & 1.00 & 1.00 & 1.00 & 1.00 & 1.00 & 1.00 & 1.00 \\
    Species \ensuremath{\to} IUCN status & 1.00 & 1.00 & 1.00 & 1.00 & 1.00 & 1.00 & 1.00 & 1.00 & 1.00 & 1.00 & 1.00 \\
    Place \ensuremath{\to} protected area & 1.00 & 1.00 & 1.00 & 1.00 & 1.00 & 1.00 & 1.00 & 1.00 & 1.00 & 1.00 & 1.00 \\
    Person \ensuremath{\to} position held & 1.00 & 1.00 & 1.00 & 1.00 & 1.00 & 1.00 & 1.00 & 1.00 & 1.00 & 1.00 & 1.00 \\
    Watercourse \ensuremath{\to} mouth & 1.00 & 1.00 & 1.00 & 1.00 & 1.00 & 1.00 & 1.00 & 1.00 & 1.00 & 1.00 & 1.00 \\
    Work \ensuremath{\to} country of publication & 1.00 & 1.00 & 1.00 & 1.00 & 1.00 & 1.00 & 1.00 & 1.00 & 1.00 & 1.00 & 1.00 \\
    Product \ensuremath{\to} manufacturer & 1.00 & 1.00 & 1.00 & 1.00 & 1.00 & 1.00 & 1.00 & 1.00 & 1.00 & 1.00 & 1.00 \\
    Person \ensuremath{\to} organization & 1.00 & 1.00 & 1.00 & 1.00 & 1.00 & 1.00 & 1.00 & 1.00 & 1.00 & 1.00 & 1.00 \\
    Work \ensuremath{\to} form & 1.00 & 1.00 & 1.00 & 1.00 & 1.00 & 1.00 & 1.00 & 1.00 & 1.00 & 1.00 & 1.00 \\
    Person \ensuremath{\to} country of birth & 1.00 & 1.00 & 1.00 & 1.00 & 1.00 & 1.00 & 1.00 & 1.00 & 1.00 & 1.00 & 1.00 \\
    Place \ensuremath{\to} body of water & 1.00 & 1.00 & 1.00 & 1.00 & 1.00 & 1.00 & 1.00 & 1.00 & 1.00 & 1.00 & 1.00 \\
    Station \ensuremath{\to} railway line & 1.00 & 1.00 & 1.00 & 1.00 & 1.00 & 1.00 & 1.00 & 1.00 & 1.00 & 1.00 & 1.00 \\
    Species \ensuremath{\to} endemic range & 1.00 & 1.00 & 1.00 & 1.00 & 1.00 & 1.00 & 1.00 & 1.00 & 1.00 & 1.00 & 1.00 \\
    Celestial object \ensuremath{\to} discoverer & 1.00 & 1.00 & 1.00 & 1.00 & 1.00 & 1.00 & 1.00 & 1.00 & 1.00 & 1.00 & 1.00 \\
    Peak \ensuremath{\to} mountain range & 1.00 & 1.00 & 1.00 & 1.00 & 1.00 & 1.00 & 1.00 & 1.00 & 1.00 & 1.00 & 1.00 \\
    Building \ensuremath{\to} street & 1.00 & 1.00 & 1.00 & 1.00 & 1.00 & 1.00 & 1.00 & 1.00 & 1.00 & 1.00 & 1.00 \\
    Work \ensuremath{\to} country of origin & 1.00 & 1.00 & 1.00 & 1.00 & 1.00 & 1.00 & 1.00 & 1.00 & 1.00 & 1.00 & 1.00 \\
    Person \ensuremath{\to} country of death & 1.00 & 1.00 & 1.00 & 1.00 & 1.00 & 1.00 & 1.00 & 1.00 & 1.00 & 1.00 & 1.00 \\
    Software \ensuremath{\to} developer & 1.00 & 1.00 & 1.00 & 1.00 & 1.00 & 1.00 & 1.00 & 1.00 & 1.00 & 1.00 & 1.00 \\
    Person \ensuremath{\to} field of work & 1.00 & 1.00 & 1.00 & 1.00 & 1.00 & 1.00 & 1.00 & 1.00 & 1.00 & 1.00 & 1.00 \\
    Book \ensuremath{\to} author & 1.00 & 1.00 & 1.00 & 1.00 & 1.00 & 1.00 & 1.00 & 1.00 & 1.00 & 1.00 & 1.00 \\
    Taxon \ensuremath{\to} parent taxon & 1.00 & 1.00 & 1.00 & 1.00 & 1.00 & 1.00 & 1.00 & 1.00 & 1.00 & 1.00 & 1.00 \\
    Cleric or abbey \ensuremath{\to} religious order & 1.00 & 1.00 & 1.00 & 1.00 & 1.00 & 1.00 & 1.00 & 1.00 & 1.00 & 1.00 & 1.00 \\
    Film \ensuremath{\to} screenwriter & 1.00 & 1.00 & 1.00 & 1.00 & 1.00 & 1.00 & 1.00 & 1.00 & 1.00 & 1.00 & 1.00 \\
    Film \ensuremath{\to} original language & 1.00 & 1.00 & 1.00 & 1.00 & 1.00 & 1.00 & 1.00 & 1.00 & 1.00 & 1.00 & 1.00 \\
    Person \ensuremath{\to} occupation (BEAR) & 1.00 & 1.00 & 1.00 & 1.00 & 1.00 & 1.00 & 1.00 & 1.00 & 1.00 & 1.00 & 1.00 \\
    Person \ensuremath{\to} alma mater & 1.00 & 1.00 & 1.00 & 1.00 & 1.00 & 1.00 & 1.00 & 1.00 & 1.00 & 1.00 & 1.00 \\
    Place \ensuremath{\to} historic county & 1.00 & 1.00 & 1.00 & 1.00 & 1.00 & 1.00 & 1.00 & 1.00 & 1.00 & 1.00 & 1.00 \\
    Work or character \ensuremath{\to} creator & 1.00 & 1.00 & 1.00 & 1.00 & 1.00 & 1.00 & 1.00 & 1.00 & 1.00 & 1.00 & 1.00 \\
    Musician \ensuremath{\to} instrument & 1.00 & 1.00 & 1.00 & 1.00 & 1.00 & 1.00 & 1.00 & 1.00 & 1.00 & 1.00 & 1.00 \\
    Celestial object \ensuremath{\to} constellation & 1.00 & 1.00 & 1.00 & 1.00 & 1.00 & 1.00 & 1.00 & 1.00 & 1.00 & 1.00 & 1.00 \\
    TV show \ensuremath{\to} network & 1.00 & 1.00 & 1.00 & 1.00 & 1.00 & 1.00 & 1.00 & 1.00 & 1.00 & 1.00 & 1.00 \\
    Person \ensuremath{\to} employer & 1.00 & 1.00 & 1.00 & 1.00 & 1.00 & 1.00 & 1.00 & 1.00 & 1.00 & 1.00 & 1.00 \\
    Artwork \ensuremath{\to} holding institution & 1.00 & 1.00 & 1.00 & 1.00 & 1.00 & 1.00 & 1.00 & 1.00 & 1.00 & 1.00 & 1.00 \\
    Athlete \ensuremath{\to} sport & 1.00 & 1.00 & 1.00 & 1.00 & 1.00 & 1.00 & 1.00 & 1.00 & 1.00 & 1.00 & 1.00 \\
    Musical work \ensuremath{\to} composer & 1.00 & 1.00 & 1.00 & 1.00 & 1.00 & 1.00 & 1.00 & 1.00 & 1.00 & 1.00 & 1.00 \\
    Athlete \ensuremath{\to} playing position & 1.00 & 1.00 & 1.00 & 1.00 & 1.00 & 1.00 & 1.00 & 1.00 & 1.00 & 1.00 & 1.00 \\
    Film \ensuremath{\to} cinematographer & 1.00 & 1.00 & 1.00 & 1.00 & 1.00 & 1.00 & 1.00 & 1.00 & 1.00 & 1.00 & 1.00 \\
    Film \ensuremath{\to} director & 1.00 & 1.00 & 1.00 & 1.00 & 1.00 & 1.00 & 1.00 & 1.00 & 1.00 & 1.00 & 1.00 \\
}
\newcommand{\finetuneAccMin}{1.00}
\newcommand{\finetuneAccMedian}{1.00}
\newcommand{\finetuneAccMean}{1.00}
\newcommand{\finetuneAccBelowOne}{0}

\begin{table}[p]
  \centering
  \caption{Accuracy on the fine-tuning subjects $\fineTuningSubjectSet$ after
    \hyperref[par:finetune]{fine-tuning on the \newrel{}}, per relation (row) and model
    (column): \candacc{} under the four \fttemplates{} after the last epoch, averaged over the
    templates within each subject, then within each answer, then over answers. Relations are
    grouped by set and sorted within it by mean \newrelgen{} over the models, highest first.
    Every entry is \finetuneAccMin{}.}
  \label{tab:finetune-acc}
  \vspace{2pt}
  \scriptsize
  \setlength{\tabcolsep}{3.5pt}\renewcommand{\arraystretch}{0.81}
  \newcommand{\finetuneAccHead}[1]{\rotatebox{90}{#1}}
  \begin{tabular}{@{}l*{\nRosterModels}{r}@{}}
    \toprule
    Relation & \finetuneAccModelHeads \\
    \midrule
\finetuneAccRows
    \bottomrule
  \end{tabular}
\end{table}

\FloatBarrier
\section{Per-relation results}
\label{app:per-relation}
\Cref{fig:per-relation-strips} shows \termGainShort{}, \newrelgen{} and \termRecallShort{} for each
of the \nPooledRelations{} relations, summarized over the \nRosterModelsWord{} models.

\FloatBarrier
\begin{figure}[p]
  \centering
  \includegraphics[width=\textwidth]{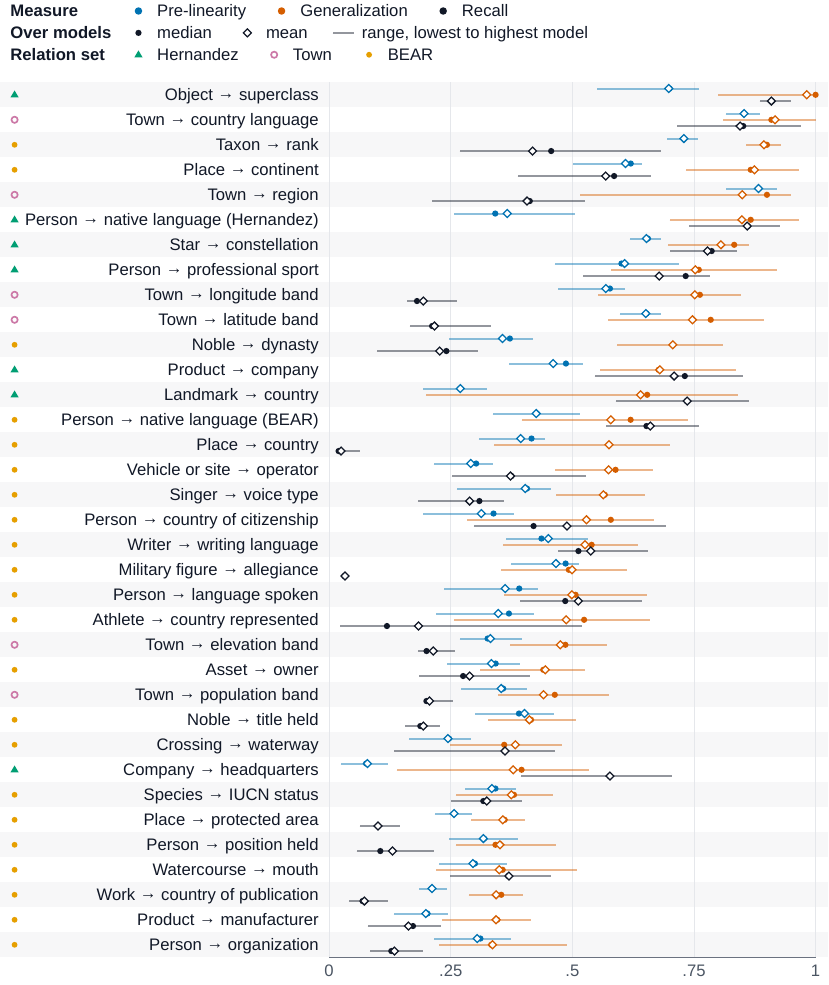}
  \caption{\textbf{\termGainShort{}, \newrelgen{} and \termRecallShort{} of every relation.}
    Relations are ordered by mean \newrelgen{} across models: the upper half of the order here,
    the lower half on the next page. Blue: \termGainShort{}; orange: \newrelgen{}; black: the
    pretrained model's \termRecallPlain{}. For each, the dot is the median over the
    \nRosterModelsWord{} models, the open diamond the mean, and the line the range (not a
    confidence interval). The mark before each name gives the relation's set: a triangle for
    \RelSetNameHernandez{}, a ring for \RelSetNameEngineered{} and a dot for \RelSetNameBear{}.}
  \label{fig:per-relation-strips}
\end{figure}
\begin{figure}[p]
  \ContinuedFloat
  \centering
  \includegraphics[width=\textwidth]{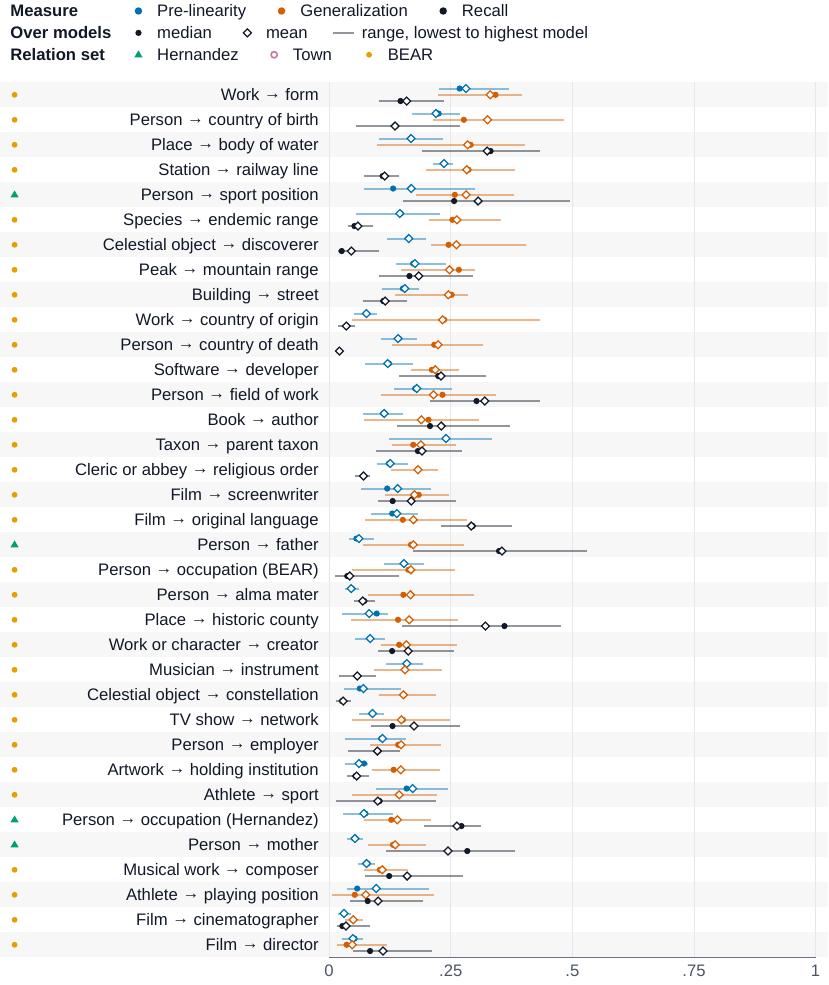}
  \caption*{\figurename~\thefigure{} (continued): the lower half of the
    relation order, with the same encoding.}
\end{figure}
\FloatBarrier

\begin{figure}[tbp]
  \centering
  \includegraphics[width=\textwidth]{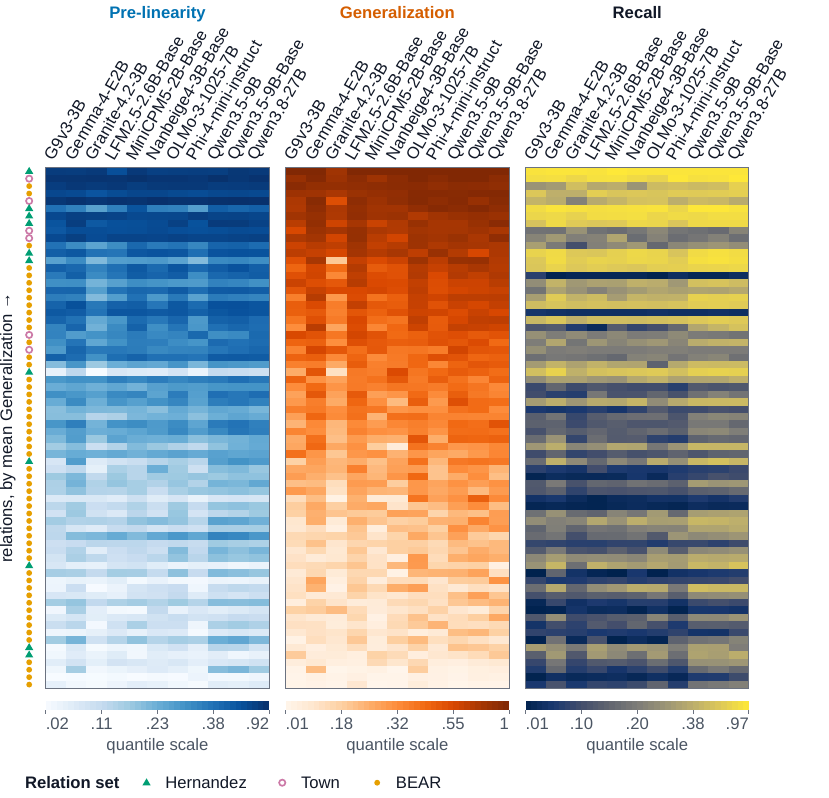}
  \caption{\textbf{\termGainShort{}, \newrelgen{} and \termRecallShort{} of every relation on
    every model.} Each row is one relation, ordered by its mean \newrelgen{} across models; each
    column is one model. The left block shows \termGainShort{}, the middle block \newrelgen{} and
    the right block the pretrained model's \termRecallPlain{}. Within each block, a square's color
    is its rank among the block's squares (a quantile scale), so every color step holds the same
    number of squares; the ticks under each color bar give the minimum, the quartiles and the
    maximum. The mark before each row gives the relation's set: a triangle for
    \RelSetNameHernandez{}, a ring for \RelSetNameEngineered{} and a dot for \RelSetNameBear{}.}
  \label{fig:relation-heatmap}
\end{figure}

\FloatBarrier
\newcommand{\repeatCorrHead}{Model & Repeat 0 & Repeat 1 & Repeat 2 & Mean of three runs \\}
\section{Correlation per repeat}
\label{app:repeat-correlation}
Tables~\ref{tab:repeat-pearson} and~\ref{tab:repeat-spearman} give the correlation of
Table~\ref{tab:cls-controls} for each of the three fine-tuning runs, on the \repeatCorrSubsetRelations{}
\RelSetNameHernandez{} and \RelSetNameEngineered{} relations for the \repeatCorrSubsetModelsWord{}
models fine-tuned three times on them. Table~\ref{tab:repeat-minicpm} gives it for
\MiniCpmModel{} on all \repeatCorrAllRelations{} relations.
\begin{table}[htbp]
  \centering
  \caption{As \cref{tab:repeat-minicpm} (Pearson $r$), for every model fine-tuned three times,
    across the \repeatCorrSubsetRelations{} \RelSetNameHernandez{} and \RelSetNameEngineered{}
    relations. Mean of three: \termGainShort{} against each relation's \newrelgen{} averaged over the three runs. $\pm$: 95\% interval (\cref{app:intervals}).}
  \label{tab:repeat-pearson}
  \vspace{2pt}
  \small
  \setlength{\tabcolsep}{5pt}\renewcommand{\arraystretch}{0.95}
  \begin{tabular}{@{}lcccc@{}}
    \toprule
    \repeatCorrHead
    \midrule
\repeatPearsonRows
    \bottomrule
  \end{tabular}
\end{table}
\begin{table}[htbp]
  \centering
  \caption{As \cref{tab:repeat-pearson}, with Spearman $\rho$ in place of Pearson $r$.}
  \label{tab:repeat-spearman}
  \vspace{2pt}
  \small
  \setlength{\tabcolsep}{5pt}\renewcommand{\arraystretch}{0.95}
  \begin{tabular}{@{}lcccc@{}}
    \toprule
    \repeatCorrHead
    \midrule
\repeatSpearmanRows
    \bottomrule
  \end{tabular}
\end{table}

\FloatBarrier
\section{\RelSetNameEngineered{} relations}
\label{app:town-relations}

The \nEngineeredRelationsWord{} \RelSetNameEngineered{} relations share one pool of 300
towns, 15 from each of 20 regions, each a real place with an English Wikipedia article. A
fixed rule gives each town its answer; a town the rule cannot
resolve has no answer in that relation and is left out of it.

Five relations read Wikidata. We match each town to a Wikidata item by its English
Wikipedia title, falling back to its name. For 50 towns the match fails: 38 find no item
and 12 find items with conflicting coordinates. These 50 towns have no answer in any of the
five. Where Wikidata gives a town several populations or elevations, we take the preferred
one, else the one with the latest point-in-time qualifier; a town whose values leave no
single choice has no answer.

Four relations cut a numeric attribute into five \emph{bands}. The bands are contiguous and
hold roughly equal numbers of towns, and towns with equal values share a band. A band with
fewer than 20 towns would be merged into its smaller neighbor; none is. A band's answer
names its position (lowest, lower, middle, upper or highest) and the range of its towns'
values, as in ``lowest population (150 to 39,653).''

\begin{itemize}
  \item \emph{Town $\to$ region}: the region we chose the town from, such as ``Borneo''
  or ``Sahara.'' 20 answers, 300 towns. New Guinea is named ``Papua'' and the South
  American rainforest ``Amazonia,'' since otherwise they would share their first token
  with New Zealand and South Africa.
  \item \emph{Town $\to$ latitude band}: latitude in degrees. 5 answers, 250 towns, 50 per
  band. The bands span $-90.0^\circ$ to $-20.7^\circ$, $-20.7^\circ$ to $-3.3^\circ$,
  $-3.1^\circ$ to $25.9^\circ$, $26.5^\circ$ to $50.1^\circ$ and $50.2^\circ$ to $78.0^\circ$.
  \item \emph{Town $\to$ longitude band}: longitude in degrees, positive to the east.
  5 answers, 250 towns, 50 per band. The bands span $-97.3^\circ$ to $-53.5^\circ$,
  $-52.9^\circ$ to $16.6^\circ$, $16.9^\circ$ to $37.7^\circ$, $39.6^\circ$ to $134.1^\circ$
  and $134.2^\circ$ to $178.0^\circ$.
  \item \emph{Town $\to$ elevation band}: elevation above sea level in meters, with values
  given in feet converted. 5 answers, 195 towns, 39 per band. Besides the 50 unmatched
  towns, 43 have no elevation and 12 have no single value. The bands span $-19$ to 9\,m,
  10 to 48\,m, 50 to 252\,m, 260 to 830\,m and 914.4 to 3830\,m.
  \item \emph{Town $\to$ population band}: number of inhabitants. 5 answers, 235 towns,
  with 48, 46, 47, 47 and 47 per band from lowest to highest. Besides the 50 unmatched
  towns, 11 have no population and 4 have no single value. The bands span 150 to 39,653,
  42,200 to 133,024, 136,302 to 297,497, 307,990 to 741,651 and 755,732 to 20,937,757.
  \item \emph{Town $\to$ country language}: the official language of the town's country,
  named in English, such as ``Japanese.'' We keep only countries with exactly one official
  language. For a country with several, such as New Zealand (English, M\=aori and New
  Zealand Sign Language), any choice would be arbitrary and could score a correct answer
  as wrong. This leaves 107 towns without an answer, besides the 50 unmatched towns and 9
  whose item names no country. The remaining
  134 towns have 17 languages. We drop the six with fewer than five towns
  (\cref{sec:setup-prerels}), which keeps 11 answers and 123 towns: Danish, English, French,
  Greenlandic, Indonesian, Japanese, Malay, Polish, Portuguese, Putonghua and Swedish.
\end{itemize}

Under every model's tokenizer, the answers of each relation have distinct first tokens, so
no two answers merge.

\FloatBarrier
\section{Label sets}
\label{app:transfer-label-pools}

\paragraph{The label-set algorithm.}
A name of the category is kept only if
(i) its primary sense lies under the category's root synset
(\texttt{animal.n.01} for animals), so the \newrelAdj{} prompt's request for
``one animal'' is true of it;
(ii) no other kept name is its hyponym, so each prompt has one correct label;
(iii) it is common in English, with \texttt{wordfreq} zipf frequency at least
\poolZipfFloor{};
(iv) its first token differs from every other kept name's under the tokenizer
of every model the label set is used with, since we score only first
tokens; and
(v) the cosine between its first token's unembedding row and any other kept
name's is at most \poolCosineCeiling{} under \MiniCpmModel{}.
For each \prerel{} $R$, the draw of $L$ then excludes every name that equals a
subject or answer of $R$, or whose primary sense lies within
\pruneStepsWord{} WordNet hypernym steps of one, so that no label is a
taxonomic neighbor of $R$'s subjects or answers (\emph{\neighbourPruning{}}).

\Cref{tab:label-pool-chain} gives the number of animal labels left after each
step of this algorithm, and \cref{tab:label-pools} the
category label sets of the label-set comparison.
The label sets, and for each relation the labels that \neighbourPruning{}
removes from each of them, are in the supplementary material under
\texttt{artifacts\_v1/label-sets/}.

\newcommand{\labelPoolChainRows}{%
    0 & Enumerate: every lemma string of \texttt{noun.animal} matching \texttt{\textasciicircum[a-z]+\$}, deduplicated & WordNet~3.0 & 3{,}048 \\
    1 & Leaves: drop a label that is a WordNet noun ancestor of another label in the label set & -- & 2{,}161 \\
    2 & Primary WordNet sense is \texttt{animal.n.01} or a hyponym of it & WordNet~3.0 sense order & 1{,}422 \\
    3 & Not named by a published profanity list & LDNOOBW English, commit \texttt{4638b97} & 1{,}422 \\
    4 & Frequency floor: \texttt{wordfreq} zipf $\geq 2.32$ & \texttt{wordfreq}~3.1.1 (\texttt{large}) & 243 \\
    5 & First token distinct from every kept label's, under all nine tokenizers at once & tokenizer revisions of record & 185 \\
    6 & Near-duplicate: first-token \texttt{lm\_head} cosine to a kept label at most 0.446 & \MiniCpmModel{} & 181 \\
    7 & Offset: in every \newrelAdj{} prompt, the label's first token is the token the trainer reads, under every roster tokenizer & checked when each run is built & 179 \\
}

\begin{table}[t]
  \centering
  \caption{The animal label set $\labelPool$, one row per step of the
    \hyperref[par:label-sets]{label-set algorithm} in the order applied, with the external
    artifact each threshold is pinned to and the labels left after the step. Every model uses
    this set: each of its \productionPoolSize{} labels has its own first token under every
    model's tokenizer, and that token is the one trained and scored.}
  \label{tab:label-pool-chain}
  \vspace{5pt}
  \small
  \setlength{\tabcolsep}{4pt}
  \resizebox{\textwidth}{!}{%
  \begin{tabular}{@{}rllr@{}}
    \toprule
    & Rule & Pinned to & Labels left \\
    \midrule
\labelPoolChainRows
    \bottomrule
  \end{tabular}%
  }
\end{table}

\newcommand{\labelPoolsRows}{%
    \texttt{animal} & \texttt{animal.n.01} & 179 & yes & 2.87 & .145 \\
    \midrule
    \texttt{act} & \texttt{act.n.02} & 460 & yes & 3.44 & .173 \\
    \texttt{artifact} & \texttt{artifact.n.01} & 393 & yes & 3.35 & .164 \\
    \texttt{attribute} & \texttt{attribute.n.02} & 245 & yes & 3.43 & .185 \\
    \texttt{body} & \texttt{body.n.01} & 0 & no & -- & -- \\
    \texttt{cognition} & \texttt{cognition.n.01} & 197 & yes & 3.45 & .169 \\
    \texttt{communication} & \texttt{communication.n.02} & 312 & yes & 3.58 & .174 \\
    \texttt{event} & \texttt{event.n.01} & 209 & yes & 3.51 & .194 \\
    \texttt{feeling} & \texttt{feeling.n.01} & 67 & no & 3.28 & .262 \\
    \texttt{food} & \texttt{food.n.01} & 57 & no & 3.49 & .210 \\
    \texttt{group} & \texttt{group.n.01} & 167 & yes & 3.71 & .178 \\
    \texttt{location} & \texttt{location.n.01} & 95 & yes & 3.82 & .188 \\
    \texttt{motive} & \texttt{motivation.n.01} & 4 & no & 4.13 & .052 \\
    \texttt{object} & \texttt{object.n.01} & 91 & yes & 3.64 & .181 \\
    \texttt{person} & \texttt{person.n.01} & 372 & yes & 3.65 & .169 \\
    \texttt{phenomenon} & \texttt{phenomenon.n.01} & 56 & no & 3.42 & .195 \\
    \texttt{plant} & \texttt{plant.n.02} & 62 & no & 3.42 & .238 \\
    \texttt{possession} & \texttt{possession.n.02} & 57 & no & 3.59 & .164 \\
    \texttt{process} & \texttt{process.n.06} & 83 & no & 3.22 & .205 \\
    \texttt{quantity} & \texttt{measure.n.02} & 151 & yes & 3.59 & .188 \\
    \texttt{relation} & \texttt{relation.n.01} & 48 & no & 3.59 & .198 \\
    \texttt{shape} & \texttt{shape.n.02} & 52 & no & 3.29 & .193 \\
    \texttt{state} & \texttt{state.n.02} & 268 & yes & 3.37 & .181 \\
    \texttt{substance} & \texttt{substance.n.01} & 233 & yes & 3.13 & .205 \\
    \texttt{time} & \texttt{time.n.05} & 8 & no & 3.94 & .232 \\
}

\begin{table}[t]
  \centering
  \caption{The label sets of the \hyperref[sec:setup-label-sets]{label-set comparison}, below
    the animal label set (\cref{tab:label-pool-chain}). Each comes from one WordNet~3.0 noun
    lexicographer file under the same algorithm, but requires every label to be a single token
    under \MiniCpmModel{} instead of a distinct first token under every model. Root: the synset
    a label's primary sense must fall under. $\geq \poolEligibilityFloor$: enough labels to take
    part. zipf med.: median \texttt{wordfreq} zipf frequency of the labels. NN cos: mean
    first-token nearest-neighbor cosine within a random 60-label draw under \MiniCpmModel{}.}
  \label{tab:label-pools}
  \vspace{5pt}
  \small
  \begin{tabular}{llrcrr}
    \toprule
    Label set & Root & $\setSize{\labelPool}$ & $\geq \poolEligibilityFloor$ & zipf med. & NN cos \\
    \midrule
\labelPoolsRows
    \bottomrule
  \end{tabular}
\end{table}

\FloatBarrier
\section{Attributing \newrelgen{} failures}
\label{app:failure-attribution}

\begin{figure}[tbp]
  \centering
  \includegraphics[width=\textwidth]{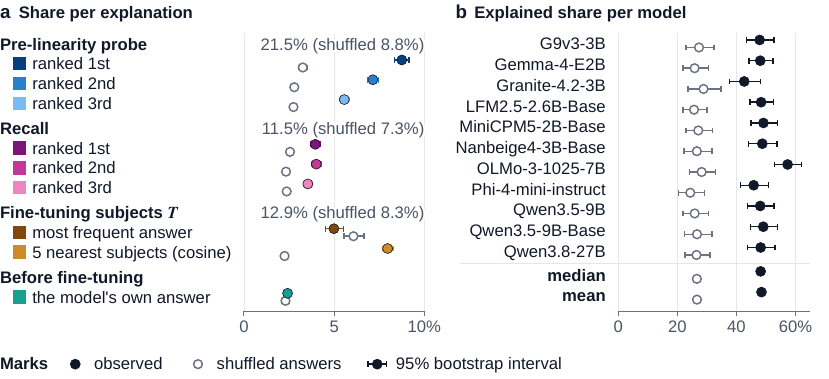}
  \caption{\textbf{Every model's errors are explained well above the shuffled baseline.}
    (a)~Each explanation's observed share of all errors against its shuffled share, grouped by
    family; a family's header gives its total. (b)~The share of each model's errors that some
    explanation accounts for, observed against shuffled, then the median and mean over models.
    Bars: 95\% intervals (\cref{app:intervals}).}
  \label{fig:failure-shapley-appendix}
\end{figure}

\paragraph{Errors.}
Fine-tuning each of the \nRosterModelsWord{} pretrained models on each of its \nPooledRelations{}
\newrel{}s gives one fine-tuned model per pair. We analyze every wrong answer the fine-tuned models
give on held-out subjects at the end of fine-tuning: \failErrors{} errors from the \failCells{}
fine-tuned models that make any; the others answer every held-out subject correctly. Through the
bijection $\pi$, each wrong label stands for a wrong answer of the \prerel{}.

\paragraph{Explanations.}
Each explanation points at one answer the model might give instead of the correct one, for example
the answer the probe ranks second. An explanation accounts for an error when the model gave exactly
that answer. If an explanation points at the correct answer, it cannot account for the error.
\begin{itemize}
  \item \emph{\termGainShort{} probe, ranked 1st, 2nd or 3rd:} the answer with the highest,
  second- or third-highest probability under the \hyperref[par:prelinearity]{probe}. The probe is
  refit on all of $\fineTuningSubjectSet$ at its most often selected layer and applied to the
  subject's last-token state.
  \item \emph{\termRecallShort{}, ranked 1st, 2nd or 3rd:} the answer the pretrained model scores
  highest, second or third highest for the subject, among the relation's answers, under the
  \prerel{}'s template.
  \item \emph{Most frequent answer:} the most frequent answer among the fine-tuning subjects
  $\fineTuningSubjectSet$.
  \item \emph{5 nearest subjects (cosine):} the majority answer of the five subjects of
  $\fineTuningSubjectSet$ nearest to the subject by cosine similarity. The similarity is computed
  between last-token states at the probe's layer, centered on the mean over
  $\fineTuningSubjectSet$; ties go to the nearer subject's answer.
  \item \emph{Answer before fine-tuning:} the model's answer on the subject's \newrel{} prompt
  before fine-tuning.
\end{itemize}

\paragraph{Shapley split.}
Several explanations can point at the same wrong answer. The error is then shared: each explanation
that accounts for it gets an equal part. This is the Shapley value of the game in which a set of
explanations is worth the number of errors that at least one of them accounts for. An error that no
explanation accounts for is unexplained. An explanation's share is its credit divided by the number
of errors, pooled over all fine-tuned models, so the shares and the unexplained rest sum to 100\%.

\paragraph{Baselines.}
An explanation can match a wrong answer by coincidence, and the two baselines measure how often. The
shuffled baseline takes the wrong answers one fine-tuned model gave and hands them out again among
its errors at random. In expectation, each error receives each of those answers, its own included,
with equal probability. An answer equal to the error's correct answer is skipped, since it would not
be an error. This keeps the model's preference for some wrong labels and breaks only their link to
the subject. The chance baseline instead gives each error a uniformly random wrong label: any label
of its \newrel{} except the correct one. Both baselines are computed exactly as expectations, not by
sampling.

The most frequent answer is the one explanation the shuffle cannot test. It points at the same
answer for every held-out subject of a fine-tuned model, so the number of wrong answers that equal
it does not depend on which subject gave which answer. Shuffling leaves that number almost
unchanged. Its shuffled share even rises, to \failShareMostFrequentShuffled{}\% from
\failShareMostFrequentObserved{}\% observed, because the other explanations match less often after
the shuffle and leave it a larger part of the credit. Against chance
(\failShareMostFrequentChance{}\%), the most frequent answer does account for more errors than
coincidence.

Intervals are 95\% percentile intervals over bootstrap resamples of the \failCells{} fine-tuned
models.

\FloatBarrier
\section{Confidence intervals}
\label{app:intervals}
Every interval is a 95\% percentile bootstrap interval: we draw 10{,}000 resamples with replacement,
recompute the statistic on each, and take the 2.5th and 97.5th percentiles. The resampled unit is what
the statistic is computed over.

\paragraph{Correlations across relations.}
For the correlations between \termGainShort{}, \termRecallShort{} and \newrelgen{}
(\cref{tab:cls-controls,tab:repeat-minicpm,tab:repeat-pearson,tab:repeat-spearman},
\cref{fig:predictor-dots,fig:label-set}), the unit is the relation: each model's coefficients are
recomputed on each resample of its relations. A
partial correlation is recomputed from scratch on each resample, with its regressions on the controls
refit and, for $\rho$, the ranks retaken. All coefficients of one model in a table or figure share the
same resamples, and the label-set comparison applies each resample to every label set. In
\cref{tab:repeat-pearson,tab:repeat-spearman}, one resample serves every model, and the median and
mean over models are recomputed from the models' resampled coefficients.

\paragraph{Error shares.}
For the Shapley shares and their baselines (\cref{fig:failure-shapley-appendix}), the unit is the
fine-tuned model: each resample draws the \failCells{} fine-tuned models with
replacement and pools their errors. A model's own interval resamples only its fine-tuned models.

\paragraph{$\pm$.}
Tables print an interval as $\pm h$, where $h$ is the larger distance from the estimate to an end of
the interval, rounded to the nearest value.

\FloatBarrier
\section{Supplementary material and licenses}
\label{app:supplementary}
The supplementary material is in our repository,
\url{https://github.com/NightMachinery/active-inert-facts}. Its directory \texttt{artifacts\_v1/} contains
the label sets (\texttt{label-sets/}), every prompt template
(\texttt{prompts/}), and every relation and \newrel{} as run
(\texttt{relations/} and \texttt{new-relations/}, and
\texttt{new-relations-label-sets/} for the label-set comparison), the
leaderboard behind the choice of models (\texttt{models/}), and our{} code
(\texttt{code/}).

The \RelSetNameBear{} relations are adapted from BEAR-big by Jacek Wiland, Max
Ploner and Alan Akbik \citep{wiland2024bear}, released under CC BY-SA 4.0
(\url{https://creativecommons.org/licenses/by-sa/4.0/}).
We keep a subset of its relations and rows and edit no answer.
The adapted rows, in this paper and in the supplementary material, are
released under the same license.
The label sets are lemmas of WordNet 3.0 \citep{miller1995wordnet}, copyright
2006 Princeton University, used under the WordNet 3.0 license.
The \RelSetNameHernandez{} relations are from the relation dataset of
\citet{hernandez2024linearity}, copyright 2022 Evan Hernandez, released under
the MIT license; the supplementary material includes its license notice.

\FloatBarrier
\section{Selection of six tiny models}
\label{app:modelselection}

Six models come from the leaderboard of tiny open-weight models of Artificial
Analysis:
\begin{quote}
\url{https://artificialanalysis.ai/models/open-source/tiny}
\end{quote}
inspected on 18 September 2026, when we chose the models; archived in the supplementary
material.
We skipped K2 Horizon 3.7B, which our code does not support.

\typeout{get arXiv to do 4 passes: Label(s) may have changed. Rerun}
\end{document}